%% file: texmain.tex
\documentclass[review]{elsarticle}

\usepackage{hyperref}

\journal{}

\makeatletter
\def\ps@pprintTitle{%
	\let\@oddhead\@empty
	\let\@evenhead\@empty
	\def\@oddfoot{\textit{{\small Preprint version \ifx\@journal\@empty \else submitted to \@journal\fi\hfill \today}}}%
	\let\@evenfoot\@oddfoot}
\makeatother

\usepackage{numcompress}\biboptions{authoryear}

\usepackage{times}
\usepackage{graphicx} 
\usepackage{subfigure} 

\usepackage{algorithm,algorithmic}

\renewcommand{\algorithmicrequire}{\textbf{Input:}}
\renewcommand{\algorithmicensure}{\textbf{Output:}}

\usepackage[draft]{fixme}

\usepackage{amsthm}

\usepackage{hyperref}

\usepackage{hl_short}
\usepackage{amssymb}
\usepackage{tikz}
\usetikzlibrary{trees}
\usetikzlibrary{shadows}
\usetikzlibrary{backgrounds}
\usetikzlibrary{shapes,arrows,automata}
\usetikzlibrary{positioning,intersections,fit,calc}
\usetikzlibrary{bayesnet}

\usepackage{enumitem}

\usepackage{bm}
\usepackage{amsmath}
\nc{\Pa}{{\rm pa}}
\nc{\Ch}{{\rm ch}}
\nc{\Co}{{\rm co}}
\nc{\MB}{{\rm mb}}
\nc{\dataset}{\calD}

\renewcommand{\KL}{\emph{KL}}
\renewcommand{\comment}[1]{\textbf{[[  #1 ]]}}
\nc{\lb}{{\cal L}}
\nc{\llb}{{\cal \hat{L}}}
\nc{\Exp}{\mathbb{E}}
\newcommand\notype[1]{\unskip}

\usepackage{lipsum}
\newcounter{examplecounter}
\newenvironment{exmp}{\begin{quote}%
		\refstepcounter{examplecounter}%
		\textbf{Example \arabic{examplecounter}}%
		\quad
	}{%
	\end{quote}%
}

\begin{document}

\begin{frontmatter}

\title{Probabilistic Models with Deep Neural Networks}

\author[ual]{Andr\'{e}s R. Masegosa}
\ead{andresmasegosa@ual.es}

\author[idsia]{Rafael Caba\~{n}as\corref{cor}}
\ead{rcabanas@idsia.ch}
\cortext[cor]{Corresponding author}

\author[ntnu]{Helge Langseth}
\ead{helge.langseth@ntnu.no}

\author[aau]{Thomas D. Nielsen}
\ead{tdn@cs.aau.dk}

\author[ual]{Antonio Salmerón}
\ead{antonio.salmeron@ual.es}

\address[ual]{Department of Mathematics, Unversity of Almer\'{\i}a, 04120 Almer\'{\i}a, Spain}
\address[idsia]{Istituto Dalle Molle di Studi sull'Intelligenza Artificiale (IDSIA), CH-6928 Manno (Lugano), Switzerland}
\address[ntnu]{Norwegian University of Science and Technology, NO-7491 Trondheim, Norway}
\address[aau]{Aalborg University, DK-9220 Aalborg, Denmark}

%
%
%
%

\begin{abstract}
\input{abstract}
\end{abstract}


\end{frontmatter}


\input{introduction}
\input{models}

\input{dnn}
\input{deeplvm}
\input{variational}
\input{conclusions}

\section*{Acknowledgements} 
This research has been partly funded by the Spanish Ministry of Science, Innovation and Universities, 
through projects TIN2015-74368-JIN, TIN2016-77902-C3-3-P and by ERDF funds.

\bibliography{join}

\end{document}

%% file: abstract.tex
Recent advances in statistical inference have significantly expanded the toolbox of probabilistic modeling. 
Historically, probabilistic modeling has been constrained to very restricted model classes, where exact or
approximate probabilistic inference is feasible.
However, developments in variational inference, a general form of approximate probabilistic inference that
originated in statistical physics, have enabled probabilistic modeling to overcome these limitations: 
(i) Approximate probabilistic inference is now possible  over a broad class of probabilistic models containing a large number of parameters, and
(ii) scalable inference methods based on stochastic gradient descent and distributed computing engines allow
probabilistic modeling to be applied to massive data sets. 
One important practical consequence of these advances is the possibility to include deep neural networks
within probabilistic models, 
thereby capturing complex non-linear stochastic relationships between the random variables. 
These advances, in conjunction with the release of novel probabilistic modeling toolboxes, have greatly expanded the scope of applications of probabilistic models, and allowed the models to take advantage of the recent strides made  by the deep learning community. In this paper we review the main concepts, methods, and tools needed to use deep neural networks within a probabilistic modeling framework. 

%% file: introduction.tex
\section{Introduction}\label{sec:Introduction}

The seminal works  about probabilistic graphical models (PGMs)  \cite{Pearl88,lauritzen1992propagation} made probabilistic modeling an indispensable tool for dealing with 
uncertainty within many different fields, such as artificial intelligence \cite{russell2016artificial}, statistics \cite{HastieTibshiraniFriedman01}, and machine learning \cite{bishop2006pattern,murphy2012machine}. 
PGMs have been present in the literature for over 30 years and have become a well established and highly influential body of research. 
%
At the same time, the problem of computing the posterior probability over hidden quantities given the known
evidence, also known as the inference problem
\cite{Pearl88,lauritzen1992propagation,JensenNielsen07,koller2009probabilistic},  has been the corner-stone as
well as the bottleneck that defines of the feasibility and applicability of probabilistic modeling. 

In the beginning, the first proposed inference algorithms \cite{Pearl88,lauritzen1992propagation} were able to compute this posterior in an exact way by exploiting the conditional independence relationships encoded by the graphical structure of the model. 
However, the set of supported probability distributions was strongly restricted, and mainly multinomial and conditional linear Gaussian distributions were used  \cite{JensenNielsen07,koller2009probabilistic}. 
Researchers quickly realized that the high computational costs of these exact inference schemes made them inappropriate for dealing with the complex stochastic dependency structures that arise in many relevant problems \cite{koller2009probabilistic} 
and, consequently, approximate inference methods became a main research focus. 

Markov Chain Monte Carlo methods were one of the first approximate methods employed for doing inference over complex PGMs \cite{gilks1995markov,salmeron2000importance,plummer2003jags}. 
These techniques are extremely versatile and powerful, and they are able to approximate complex posterior distributions. 
However, they have serious issues wrt., e.g., the convergence of the underlying Markov chain and poor mixing when approximating high dimensional distributions \cite{gilks1995markov}. 
Computing such high dimensional posteriors started to become relevant in many domains, specifically when researchers 
applied a Bayesian approach for learning the parameters of their PGMs from data \cite{bishop2006pattern,murphy2012machine,blei2014build}. 
In this setup, the model parameters are treated as unobserved random variables, and the learning problem
therefore reduces to computing the posterior probability over the parameters. 
For models with a large number of parameters, the approach leads to  high dimensional posteriors, where the application of Monte Carlo methods becomes infeasible. 
These issues gave rise to the development of alternative approximate inference schemes. 

{Belief propagation} (BP) \cite{Pearl88,murphy1999loopy}, and the closely related {Expectation propagation}
(EP) algorithm \cite{minka2001expectation}, have been successfully used 
to overcome many of the limitations of Monte Carlo methods. 
These deterministic approximate inference techniques can be implemented using a message-passing scheme that takes advantage of the graph structure of the PGM and, hence, the underlying conditional independence relationships among variables. 
In terms of distributional assumptions, BP has mainly been used with multinomial and Gaussian distributions. Although EP allows for a more general family of distributions, it is restricted by the need to define a non-trivial quotient operation between the involved densities. 
While these techniques overcame some of the difficulties of Monte Carlo methods,  they presented two new issues: 
$(i)$ they do not guarantee convergence to an approximate and meaningful solution; 
and $(ii)$ do not scale to the kind of models that appear in the context of Bayesian learning 
\cite{murphy2012machine,blei2014build}. 
Again, these challenges motivated researchers to look into  alternative approximate inference schemes. 

Variational methods \cite{wainwright2008graphical} were firstly explored in the context of PGMs during the late 90s \cite{jordan1999introduction}, inspired by their successful application to inference problems encountered in statistical physics. 
Like BP and EP, they are deterministic approximate inference techniques. 
The main innovation is to cast the inference problem as a minimization-problem with a well defined loss function, namely the negative Evidence Lower BOund (ELBO) function, which acts as an inference proxy. 
In general, variational methods guarantee convergence to a local maximum of this ELBO function and therefore to a meaningful solution. 
By transforming the inference problem into a continuous optimization problem, variational methods can take advantage of recent advances in continuous optimization theory.
This was the case for the widely adopted stochastic gradient descent algorithm \cite{bottou2010large}, which
has successfully been used by the machine learning community to scale learning algorithms to big data sets. 
This same learning algorithm was adapted to variational inference in \cite{HoffmanBleiWangPaisley13}, giving the opportunity to apply probabilistic modeling  to problems involving massive data sets.  
In terms of distributional assumptions, these variational inference methods were restricted to the conjugate
exponential family \cite{barndorff2014information}, where the gradient of the ELBO wrt.\ the model parameters can be computed in closed-form \cite{WinnBishop05}.  Ad-hoc approaches were developed for models outside this distributional family. 

From the start of the field at end of the 1980's and up to around 2010, probabilistic models had mainly  been focused on
using distributions from the conjugate exponential family, even though this family of distributions is only able to model linear relationships between   the random variables \cite{WinnBishop05}. 
On the other hand, one of the reasons for the success of deep learning methods \cite{goodfellow2016deep} is the ability of deep neural networks to model non-linear relationships among high-dimensional objects, as is, e.g., observed between the pixels in an image or the words in a document.
Recent advances in variational inference \cite{kingma2013auto,ranganath2014black} have enabled the integration
of deep neural networks  in probabilistic models, thus also making it possible to capture such non-linear
relationships among the random variables. This gave rise to a whole new family of probabilistic models, which are often denoted \textit{deep generative models} \cite{hinton2009deep,hinton2012practical,goodfellow2014generative,salakhutdinov2015learning}.
This new family of probabilistic models are able to encode objects like images, text, audio, and video probabilistically, 
thus bringing many of the recent advances produced by the deep learning community to the field of probabilistic modeling. 
The release of modern probabilistic programming languages
\cite{tran2016edward,tran2018simple,bingham2018pyro,cabanasInferPy,cozar2019inferpy} relying on well
established deep learning engines like Tensorflow \cite{tensorflow2015-whitepaper} and PyTorch
\cite{paszke2017automatic} have also significantly contributed to the adoption of these powerful probabilistic modeling techniques. 


In this paper we give a coherent overview of the key concepts and methods needed for integrating deep neural networks in probabilistic models. 
The present paper differs from other recent reviews of, e.g., deep generative models
\cite{salakhutdinov2015learning} and variational inference methods \cite{zhang2018advances}, as we also go
into details regarding the implementation of such models using relevant software tools. 
To this end, the paper is accompanied by online material, where the running examples of the paper together
with other basic probabilistic models containing artificial neural networks are implemented to illustrate the
theoretical concepts and methods presented in the paper.\footnote{\url{https://github.com/PGM-Lab/ProbModelsDNNs}}

%% file: models.tex
\section{Probabilistic Models within the Conjugate Exponential Family}\label{sec:models}

\subsection{Latent Variable Models}\label{sec:LVM}

The conjugate exponential family of distributions \cite{barndorff2014information} 
covers a broad and widely used range of probability distributions and density functions such as Multinomial, 
Normal, Gamma, Dirichlet and Beta. 
They have been used by the machine learning community \cite{bishop2006pattern,koller2009probabilistic,murphy2012machine}  
due to their convenient properties related to parameter learning and inference tasks. 

\begin{figure}
\begin{center}
\scalebox{0.95}{
\begin{tikzpicture}
  \node[latent]  (beta)   {$\boldsymbol{\beta}$}; %
  
  \node[const, left= .5  of beta]  (alpha)   {$\bmalpha$}; %
  \edge{alpha}{beta};
  
  \node[obs, below right=1 of beta]          (X)   {$\bmX_{i}$}; %
  \edge{beta}{X};
 
  \node[latent, below left= 1 of beta]          (Z)   {$\bmZ_{i}$}; %
  \edge{Z}{X};
  \edge{beta}{Z};
  
  \plate {Observations} { %
    (X)(Z)
  } {\tiny $i={1,\ldots, N}$} ;


\end{tikzpicture}
}

\end{center}
\vspace{-3mm}
\caption{\label{fig:plateModel} Structure of the probabilistic model examined in this paper, defined for a sample of size $N$. }
\end{figure}
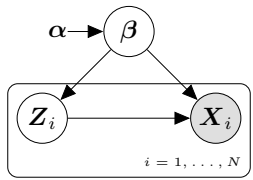

In the following we focus on probabilistic graphical models with structure as shown in \figref{plateModel}, and where the full model belongs to the conjugate exponential family. 
These 
models are also known as latent variable models (LVMs) \cite{bishop1998latent,blei2014build}. 
LVMs are widely used as a tool for discovering patterns in data sets. 
The model in \figref{plateModel} captures ``local'' patterns, which are specific to sample $i$ of the data, using unobservable (or latent) random variables denoted by $\bmZ_i$. 
``Global'' patterns, those that are shared among all the samples of the data set,  are modelled by means of a set of latent random variables denoted by $\bmbeta$.
The observed data sample $i$, $\bmX_i$, is modelled as random variables whose distribution is conditioned on both the 
local ($\bmZ_i$) and global ($\bmbeta$) latent variables. $\bmalpha$, a vector of fixed hyper-parameters,  is also included in the model. 

While the model structure in \figref{plateModel}  at first sight can appear restrictive, it is in fact quite versatile, and many books contain entire sections devoted to LVMs \cite{bishop2006pattern,koller2009probabilistic,murphy2012machine}.
For instance, LVMs include popular models like Latent Dirichlet Allocation (LDA) models used to uncover the hidden topics in a text corpora \cite{blei2003latent}, mixture of Gaussian models to discover hidden clusters in  data \cite{bishop2006pattern}, probabilistic principal component analysis for dimensionality reduction \cite{tipping1999probabilistic}, and models to capture  drift in a data stream \cite{masegosa2017bayesian}.  
They have been used for knowledge extraction from GPS data \cite{kucukelbir2017automatic}, genetic data \cite{pritchard2000inference}, graph data \cite{kipf2016variational}, and so on.



The joint distribution of this probabilistic model factorizes into a product of local terms and a global term as
\[
p(\bmx, \bmz, \bmbeta)  = p(\bmbeta) \prod_{i=1}^N p(\bmx_i, \bmz_i|\bmbeta),
\]
\noindent where $N$ is the number of samples. As can be seen, the local latent variables $\bmZ_i$ are assumed conditionally independent given the global latent variables $\bmbeta$.

Another standard assumption in these models is known as the assumption of \textit{complete conditional form}  \cite{HoffmanBleiWangPaisley13}. 
Now, the distribution of one latent variable given the the other variables in the model can be expressed in exponential family form, 
\begin{equation}\label{eq:posteriorexponentialform} 
\begin{split}
\ln p(\bmbeta|\bmx, \bmz) = h_g(\bmbeta)  + \bmeta_g(\bmx, \bmz)\trans \bmt(\bmbeta) - a_g(\bmeta_g(\bmx, \bmz)),\\
\ln p(\bmz_i|\bmx_i, \bmbeta) = h_l(\bmz_i)  + \bmeta_l(\bmx_i,\bmbeta)\trans \bmt(\bmz_i) - a_l(\bmeta_l(\bmx_i,\bmbeta)).
\end{split}
\end{equation}
\noindent where the scalar functions $h_\cdot(\cdot)$ and $a_\cdot (\cdot)$ are the base measures and the log-normalizers functions,
respectively; the vector functions $\bmeta_\cdot(\cdot)$ and $\bmt_\cdot(\cdot)$ are the \textit{natural parameter} and the \textit{sufficient statistics} vectors,  respectively. 
The subscripts of these functions, here $g$ for ``global'' and $l$ for ``local'', are used to signify that the different functions differ between variables. 
The subscripts will be removed when clear from context. 

By conjugacy properties, the above assumptions also ensure that the conditional distribution $p(\bmx_i,\bmz_i|\bmbeta)$ is in the exponential family, 
\begin{equation}\label{eq:jointexponentialform}
\ln p(\bmx_i,\bmz_i|\bmbeta) = \ln h(\bmx_i,\bmz_i) + \bmbeta^{T} \bmt(\bmx_i,\bmz_i) - a(\bmbeta),
\end{equation}
\noindent and, similarly, for the prior distribution $p(\bmbeta)$,
\begin{eqnarray}\label{eq:model}
\ln p(\bmbeta) &=& \ln h_{\bm \beta} (\bmbeta) + \bmalpha\trans \bmt_{\bm \beta}(\bmbeta) - a_{\bm \beta}(\bmalpha).
\end{eqnarray}

Combining Equation (\ref{eq:jointexponentialform}) and Equation (\ref{eq:model}), we see that the posterior $p(\bmbeta|\bmx, \bmz)$ remains in the same distribution family as the prior $p(\bmbeta)$ (that is, we have conjugacy) and, in consequence, the natural parameter of the global posterior $\bmeta_g(\bmx, \bmz)$ can be expressed as
\begin{equation}\nonumber
\bmeta_g(\bmx, \bmz) = \bmalpha + \sum_{i=1}^N \bmt(\bmx_i,\bmz_i).
\end{equation}
This representation of the \textit{complete conditional} will be used  later to derive the variational inference scheme over this model.

\begin{figure}[htb!]

	\resizebox{\linewidth}{!}{
	\begin{tabular}{cc}
	\includegraphics{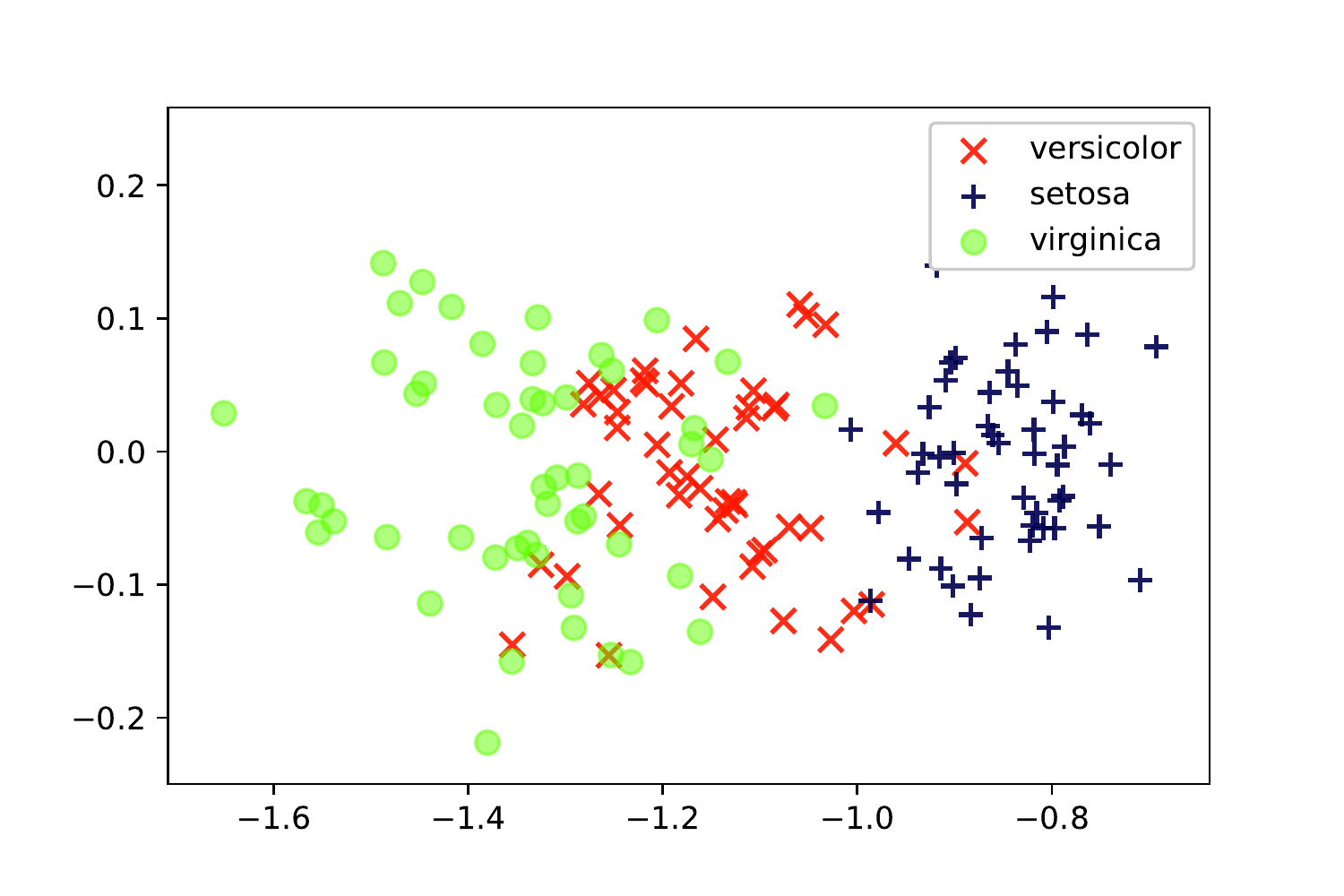}
	&
		\includegraphics{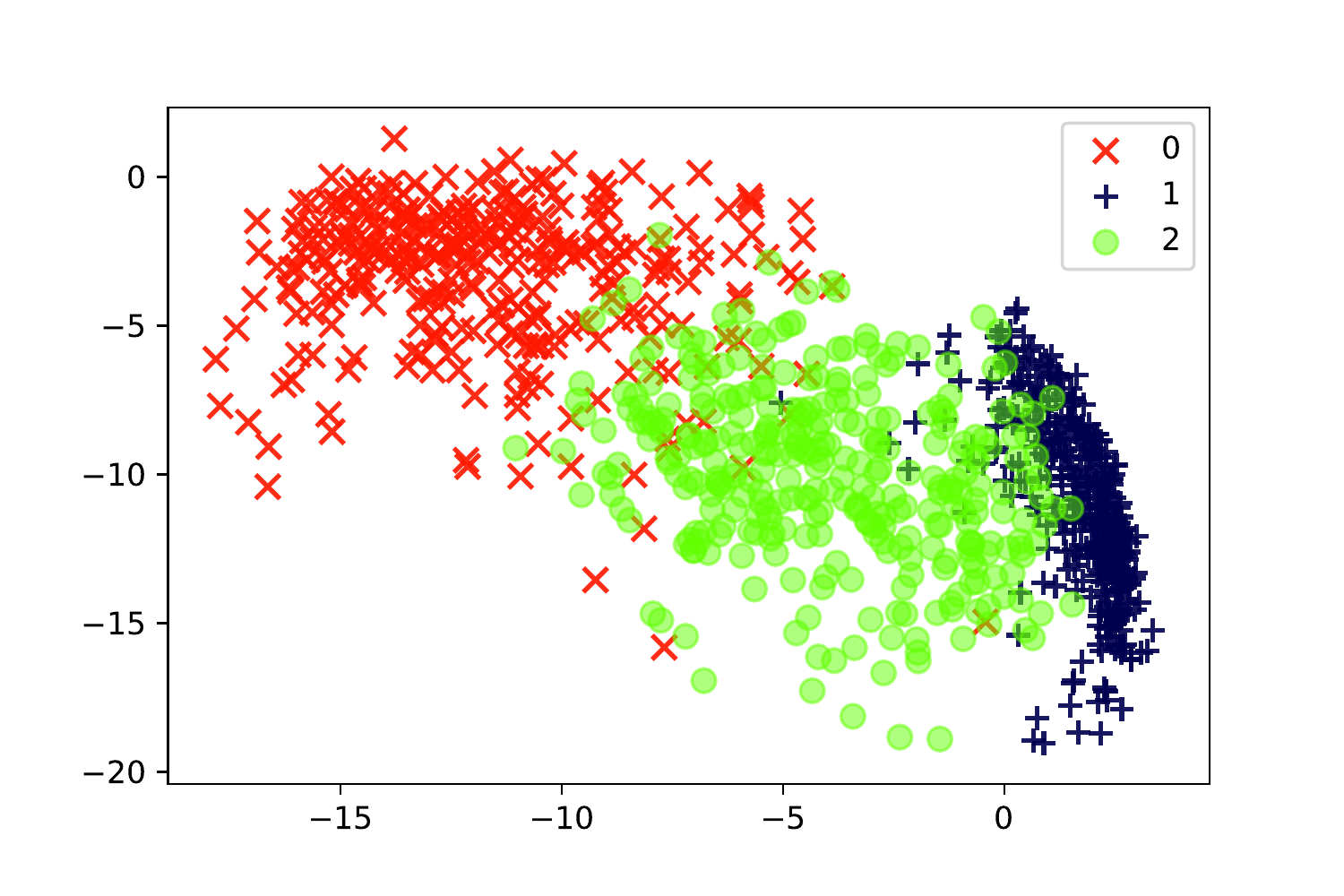}
	\end{tabular}
	}
	\caption{\label{fig:PCA:application}2-dimensional latent representations resulting of applying a probabilistic PCA of: \textbf{(Left)} the iris dataset \citep{fisher1936use} and \textbf{(Right)} a subset of 1000 instances from the MNIST dataset \citep{lecun1998gradient} corresponding to the handwritten digits 1, 2 and 3.}
\end{figure}

\begin{exmp}\label{example:PCA}
Principal Component Analysis (PCA) is a classic statistical technique for dimensionality reduction. It defines a mapping between the $d$-dimensional data-representation of a point $\bmx$ and its $k$-dimensional latent representation, $\bmz$. The latent representation is known as the \textit{scores}, and the affine transformation is performed using the \textit{loading matrix}  $\bmbeta$, which has dimensions $k\times d$. 


\begin{algorithm}[htb!]
\caption{Pseudo-code of the generative model of a probabilistic PCA model.}\label{alg:pca}
\begin{algorithmic}
\STATE \# Sample from global random variables
\STATE $\beta_{u,v} \sim {\cal N} (0, 1)$   ~~~\# Sample for $u=1\ld k$, $v=1\ld d$.
\FOR{$i=1\ld N$}
\STATE \# Sample from the local latent variables 
\STATE $\bmz_i \sim {\cal N} (\bm0,\bmI)$
\STATE \# Sample from the observed variables 
\STATE $\bmx_i \sim {\cal N} (\bmbeta\trans\bmz_i, \sigma^2_\bmx\bmI)$
\ENDFOR
\end{algorithmic}
\end{algorithm}


A simplified probabilistic view of PCA \cite{tipping1999probabilistic}  is given in Algorithm \ref{alg:pca}, which provides pseudo-code for the generative process of a probabilistic PCA model.
This model is obviously an LVM, as the loadings represented by $\bmbeta$ are global latent variables and $\bmZ_i$ is the vector of local latent variables associated with the $i$-th element in the sample. 

This model belongs to this conjugate exponential family with complete conditionals, because the joint of $p(\bmx,\bmz,\bmbeta)$ is multivariate Normal and, by standard properties of the multivariate Normal distribution,  the conditional $p(\bmbeta|\bmz,\bmx)$ and $p(\bmz_i|\bmx_i,\bmbeta)$ are both conditional multivariate Gaussians. A multivariate Normal distribution with mean $\bmmu$ and covariance matrix $\bmSigma$ is a member of the exponential family with natural parameters $\bmeta = \left[\bmSigma^{-1}\bmmu, -1/2\bmSigma^{-1}\right]\trans$ and sufficient statistics $\bmt(\bmx)=\left[\bmx,\bmx\bmx\trans\right]\trans$.


Note that while this linear relationship between the latent and the observed variables is a strong limitation of this model \cite{scholkopf1998nonlinear}, it guarantees that the model belongs to conjugate exponential family. 
Using a non-linear relationship would put PCA outside this  model family and would prevent, as we will see in the next section, the use of efficient inference algorithms to calculate $p(\bmbeta|\bmz,\bmx)$ and $p(\bmz_i|\bmx_i,\bmbeta)$. Similarly, if the variance parameter $\sigma_x$ (see Algorithm \ref{alg:pca}) depend on the latent variables $\bmz_i$, the model falls outside the conjugate exponential family.

Figure \ref{fig:PCA:application}  illustrates the behavior of Probabilistic PCA as a feature reduction method on two different data sets, \textit{Iris} and (a reduced version of) \textit{MNIST}. The data is projected from data-dimension $d=4$ (Iris) or $d=784$ (MNIST)  down into $k=2$ latent dimensions.
As can be seen, the method captures some of the underlying structure in the Iris-data, and even generates a representation where the three classes of the flower can be separated. 
On the other hand, the MNIST representation appears less informative. Images of the three digits ``1'', ``2'' and ``3'' are given to the PCA, but even though these three groups of images are quite distinct, the learned representation is not able to clearly separate the classes from one another. 
As we will see later in this paper, when we consider a more expressive mappings between the local latent $\bmZ_i$ and $\bmX_i$ (using artificial neural networks), the latent representations will become more informative.
\end{exmp}

\subsection{Mean-Field Variational Inference}\label{sec:VI}

The problem of Bayesian inference reduces to computing the posterior over the unknown quantities (i.e. the global and local latent variables $\bmbeta$ and $\bmz$, respectively) given the observations, 

\[
p(\bmbeta, \bmz\given \bmx) = \frac{p(\bmx|\bmz,\bmbeta)p(\bmz\given \bmbeta)p(\bmbeta)}{\int \int p(\bmx|\bmz,\bmbeta)p(\bmz\given \bmbeta)p(\bmbeta) d\bmz d\bmbeta}.
\]
Computing the above posterior is intractable for many interesting models, because it requires to solve the complicated multidimensional integral in the denominator. 
As commented in the introduction, variational inference (VI) methods are one of the best performing options to address this problem. 
In this section we revise the main ideas behind this approach. 

\begin{exmp}\label{example:PCAInference}
Computing $p(\bmbeta, \bmz\given \bmx)$ for the probabilistic PCA model described in Example \ref{example:PCA} is not feasible since the integral 
$$p(\bmx)=\int\!\!\int p(\bmx|\bmz,\bmbeta)p(\bmz\given \bmbeta)p(\bmbeta) d\bmz d\bmbeta$$ is intractable.  
The source of the problem is that $p(\bmx\given\bmbeta)=\int p(\bmx|\bmz,\bmbeta)p(\bmz\given \bmbeta) d\bmz$ is not in the conjugate exponential family.


\vekk{
\begin{eqnarray*}
p(\bmx) &=&\int_{\bm \beta} p_\calN(\bmbeta|\bm0,\bmI\sigma^2_{\bm \beta}) \prod_i \int_z p_\calN (\bmx_i|\bmz_i \bmbeta\trans, \sigma^2_x) p_\calN(\bmz|\bm0,\bmI) d\bmz d\bmbeta\\ 
&=& \int_{\bm \beta} p_\calN(\bmbeta|\bm0,\bmI\sigma^2_{\bm \beta}) \prod_i p_\calN (\bmx_i|\bm0, \sigma^2_x + \bmbeta\trans\bmbeta) d\bmbeta,
\end{eqnarray*}
\noindent where  $p_\calN(\cdot|\mu,\sigma^2)$ denotes the density function of a Normal distribution. The last integral of the above equation is intractable because it involves marginalizing a Normal variable $\bmbeta$ which affects the covariance matrix of another Normal random variable.
}

\end{exmp}

%
Variational inference is a deterministic technique that finds a tractable approximation to an intractable (posterior) distribution. 
We will use $q$ to denote the approximation, and use $p$ to signify the true distribution (like $p(\bmbeta,\bmz\given\bmx)$ in the example above).
More specifically, let  ${\cal Q}$ denote a set of possible approximations $q$. 
Now, VI solves the following optimization problem:
\begin{equation} \label{eq:VIKL-minimization}
\min_{q\in {\cal Q}} \KL(q||p),
\end{equation}
\noindent where $\KL$ denotes the Kullback-Leibler divergence between two probability distributions. For the specific problem at hand, this general formulation is more precisely written as 
\[
\min_{q\left(\bmbeta,\bmz\right)\in {\cal Q}} \KL(q(\bmbeta,\bmz)||p(\bmbeta,\bmz|\bmx))
\]
\noindent 
Notice that while $q$ depends on the observations $\bmx$, it is customary to make this implicit in the notation, and write, e.g., $q\left(\bmbeta, \bmz\right)$ instead of 
$q\left(\bmbeta,\bmz\given\bmx\right)$. 
In practice, one will typically posit that $\calQ$ is a convenient distributional family indexed by some parameters, say $\bmtheta$, and the minimization of Equation \eqref{eq:VIKL-minimization} amounts to finding the parameters $\bmtheta^\star$ that minimize the KL divergence.

Under the \textit{mean field} variational approach, the approximation family ${\calQ}$ is assumed to fully factorize. 
Following the notation in \cite{HoffmanBleiWangPaisley13}, we have that
\begin{equation}
\label{eq:meanfieldQ}
q(\bmbeta,\bmz\given \bmlambda,\bmphi) = q(\bmbeta \given \bmlambda)\prod_{i=1}^N  q(\bmz_i \given \bmphi_i),
\end{equation}
 
\noindent where $\bmlambda$ parameterizes the variational distribution of $\bmbeta$, while $\bmphi_i$ has the same role for the variational distribution of $\bmZ_i$.

Furthermore, if the model is model in the conjugate exponential family, each factor in the variational distribution is assumed to belong to the same family of the model's \textit{complete conditionals} (see Equation~\eqref{eq:posteriorexponentialform}), 
\begin{equation}\label{eq:qcompleteconditionals}
\begin{split}
\ln q(\bmbeta|\bmlambda) = h(\bmbeta)  + \bmlambda\trans \bmt(\bmbeta) - a(\bmlambda),\\
\ln q(\bmz_i|\bmphi_i) = h(\bmz_i)  + \bmphi_i\trans \bmt(\bmz_i) - a(\bmphi_i).
\end{split}
\end{equation}

%

To solve the minimization problem in Equation \eqref{eq:VIKL-minimization}, the variational approach exploits  the  transformation
\begin{equation}
\label{equ:likelihood_decomposition}
\ln p(\bmx) = \lb(\bmlambda,\bmphi) + \KL(q(\bmbeta,\bmz\given \bmlambda,\bmphi)||p(\bmbeta,\bmz|\bmx)),  
\end{equation}
\noindent where $\lb$  can be expressed as  
\begin{equation}\label{eq:elbo}
\lb(\bmlambda,\bmphi)  = \E_q [\ln p(\bmx, \bmZ, \bmbeta)] - \E_q [\ln q(\bmbeta,\bmZ|\bmlambda,\bmphi)].
\end{equation}
\noindent
$\lb$ is of interest in its own right. 
Notice in particular that $\lb$ in \equref{likelihood_decomposition} is a \emph{lower bound} of $\ln p(\bmx)$ since the KL-divergence is  non-negative. For this reason, $\lb$  is usually referred to as the ELBO (Evidence Lower BOund).
Furthermore, as  $\ln p(\bmx)$ is constant in the optimization wrt.\ $q$, minimizing the KL divergence in Equation \eqref{eq:VIKL-minimization} is equivalent to maximizing the lower bound $\lb$. 
Variational methods maximize $\lb$  using gradient based techniques. 

The key advantage of having a conjugate exponential model is that the gradients of  $\lb$  wrt.\ its parameters can always be computed in closed form ~\cite{WinnBishop05}. 
This is important, as it leads to a natural scheme in which the parameters are updated iteratively: For a parameter $\theta_j$, simply choose the 
value $\theta_j^\star$ so that 
$\left.\nabla_{\theta_j} \lb (\bmtheta)\right|_{\bmtheta:\theta_j=\theta_j^\star}=0$.
In practice it is beneficial to use  the \textit{natural gradients}, which is the standard gradient pre-multiplied by the inverse of the Fisher information matrix, to account for the Riemannian geometry of the parameter space \citep{amari1998natural}.

The gradients with respect to the variational parameters $\bmlambda$ and $\bmphi$ can be computed as follows, 
\begin{equation}\label{eq:gradELBO}
\begin{split}
\nabla^{nat}_{\bm \lambda} \lb (\bmlambda,\bmphi) &= \bmalpha + \sum_{i=1}^N  \E_{\bmZ_i} [\bmt(\bmx_i,\bmZ_i)]  - \bmlambda ,\\
\nabla^{nat}_{\bm \phi_i}\lb (\bmlambda,\bmphi) &=  \E_{\bmbeta} [\bmeta_l(\bmx_i,\bmbeta)]  - \bmphi_i,
\end{split}
\end{equation}
\noindent where $\nabla^{nat}$ denotes natural gradients
and $\E_{\bmZ_i}[\cdot]$ and $\E_{\bmbeta} [\cdot]$ denote expectations with respect to $q(\bmz_i\given\bmphi_i)$ and $q(\bmbeta\given\bmlambda)$, respectively.


From the above gradients we can derive a coordinate ascent algorithm to optimize the ELBO function with the following coordinate ascent rules, 
\begin{equation}\label{eq:CoordinateAscent}
\begin{split}
\bmlambda^\star &=  \arg\max_{\lambda} \lb (\bmlambda, \bmphi) = \bmalpha + \sum_{i=1}^N  \E_{\bmZ_i} [\bmt(\bmx_i,\bmZ_i)] , \\
\bmphi_i^\star &= \arg\max_{\bm \phi_i}  \lb (\bmlambda, \bmphi) = \E_{\bmbeta} [\bmeta_l(\bmx_i,\bmbeta)].
\end{split}
\end{equation}

By iteratively running the above updating equations, we are guranteed to (i) monotonically increase the ELBO function at every time step and (ii) to converge to a stationary point of the ELBO function or, equivalently, the  function minimizing Equation~\eqref{eq:VIKL-minimization}.


\begin{exmp}\label{example:PCA:VI}
For the PCA model in Example \ref{example:PCA}, the variational distributions are 
\begin{eqnarray*}
q(\bmbeta\given \bmmu_{\beta},  \bmSigma_{\beta}) &=& {\cal N}(\bmbeta\given\bmmu_{\beta},  \bmSigma_{\beta}), \\
q(\bmz_i\given \bmmu_{z_i}, \bmSigma_{z_i}) &=&  {\cal N}(\bmz_i\given\bmmu_{z_i},   \bmSigma_{z_i}).
\end{eqnarray*}  

Given the above variational family, the coordinate updating equations derived from Equation~\eqref{eq:CoordinateAscent} can be written, after some algebraic manipulations, as \cite{bishop2006pattern}
\begin{eqnarray*}
\bmSigma_{\beta} &=& \left(\sum_{i=1}^N \E[\bmZ_i\bmZ_i\trans] +  \sigma^2_x \bmA \right)^{-1}, \\
\bmmu_{\beta} &=& \left[\sum_{i=1}^N \bmx_i\E[\bmZ_i]\right]\trans\bmSigma_{\beta} , \\
\bmSigma_{z_i} &=& \left(\bmI +\bmmu_{\beta}\trans\bmmu_{\beta} / \sigma_x^{2} \right)^{-1}, \\
\bmmu_{z_i} &=& \bmSigma_{z_i}\bmmu_{\beta}\trans\bmx_i / \sigma_x^{2} , 
\end{eqnarray*}  
\noindent where $\bmA$ is a diagonal matrix with element at index $(i, i)$ given by $d/\bmmu_{\beta,i}\trans\bmmu_{\beta,i}$.
Again, we have a  set of closed-form equations which guarantees convergence to the solution of the inference problem. 
We should note that this is possible due to the strong assumptions imposed both on the probabilistic model $p$ and on the family of variational approximations $\calQ$. 


\end{exmp}



\subsection{Scalable Variational Inference}\label{sec:ScalableVI}

Performing VI on large data sets (measured by the number of samples, $N$) raises many challenges. 
Firstly, the model itself may not fit in memory, and, secondly, the cost of computing the gradient of the ELBO with respect to $\bmlambda$ linearly 
depends  on the size of the data set (see Equation~\eqref{eq:gradELBO}), which can be prohibitively expensive when $N$ is very large.  
Stochastic Variational inference (SVI) \citep{HoffmanBleiWangPaisley13} is a popular method for scaling VI to massive data sets, 
and relies on stochastic optimization techniques \citep{robbins1951stochastic,bottou2010large}. 

We start by re-parameterizing the ELBO so that $\lb$ is expressed only in terms of the global parameters $\bmlambda$. 
This is done by defining 
\begin{equation}\label{eq:scalable:elbo}
\lb(\bmlambda) = \lb(\bmlambda,\bmphi^\star(\bmlambda)) ,
\end{equation}
\noindent where $\bmphi^\star(\bmlambda)$ is defined as in Equation~\eqref{eq:CoordinateAscent}, i.e. it returns a local optimum of the local variational parameters $\bmphi$ for a given $\bmlambda$. 
Now $\lb(\bmlambda) $ has the following form:
\begin{equation}\label{eq:scalable:elboreparam}
\begin{split}
\lb(\bmlambda) &= \E_q[\ln p(\bmbeta)] - \E_q[\ln q(\bmbeta\given\bmlambda)] \\
&+ \sum_{i=1}^N \max_{\bm \phi_i}\left\{\E_q[\ln p(\bmx_i,\bmZ_i\given\bmbeta)] - \E_q[\ln q(\bmZ_i\given\bmphi_i)]  \right\}
\end{split}
\end{equation}

As shown in \cite{HoffmanBleiWangPaisley13}, we can compute the gradient of $\lb(\bmlambda)$ by first finding $\bmphi^\star(\bmlambda)$, and then compute the gradient w.r.t. $\bmlambda$ while keeping $\bmphi^\star(\bmlambda)$ fixed (because $\nabla_{\bm \lambda} \lb(\bmlambda) = \nabla_{\bm \lambda} \lb(\bmlambda,\bmphi^\star(\bmlambda))$). 
By exploiting properties of the conjugate exponential family, the natural gradient with respect to $\bmlambda$ can be computed in closed-form as
\begin{equation}\nonumber
\nabla^{nat}_{\bm \lambda} \lb(\bmlambda) = \bmalpha + \sum_{i=1}^N  
\E_{q(\bmz_i\given\phi_i^\star)}[\bmt(\bmx_i,\bmZ_i)]  - \bmlambda .\\
\end{equation}

The key idea behind SVI is to compute unbiased albeit noisy estimates of $\nabla^{nat}_{\bm \lambda} \lb$,  denoted $\hat{\nabla}^{nat}_{\bm \lambda}\lb$,
by randomly selecting a mini-batch of $M$ data samples, and then define
\begin{equation}\nonumber
\begin{split}
\hat{\nabla}^{nat}_{\bm \lambda} \lb (\bmlambda) = \bmalpha + \frac{N}{M}\sum_{m=1}^M \E_{q(\bmz_i\given\phi_i^\star)} [t (\bmx_{i_m},\bmZ_{i_m})]  - \bmlambda,
\end{split}
\end{equation}
\noindent where $i_m$ is the variable index form the subsampled mini-batch. 
It is immediate that $\E[\hat{\nabla}^{nat}_{\bm \lambda} \lb] = \nabla^{nat}_{\bm \lambda} \lb$, hence the estimator is unbiased. 
Utilizing stochastic optimization theory \cite{robbins1951stochastic}, the ELBO can be maximized by following noisy estimates of the gradient, 
\begin{equation}\label{eq:gradELBONoisy}
\begin{split}
\bmlambda_{t+1} = \bmlambda_t + \rho_t \hat{\nabla}^{nat}_{\bm \lambda} \lb(\bmlambda_t),
\end{split}
\end{equation}
\noindent if the learning rate $\rho_t$ satisfies the Robbins-Monro conditions\footnote{A sequence $\{\rho_t\}_{t=1}^\infty$ satisfies 
the Robbins-Monro conditions if $\sum_{t=1}^\infty \rho_t = \infty$ and $\sum_{t=1}^\infty \rho^2_t<\infty$.}. 
In this case the above updating equation is guaranteed to converge to a stationary point of the ELBO. 

To choose  the size of the mini-batch $M$, two conflicting issues should be considered: Smaller values of $M$ (i.e., $M \ll N$) leads to a reduction in the computational complexity of computing the gradient, while larger values of $M$ (i.e., $M\gg1$) reduces the variance of the estimator. 
The optimal value for $M$ is problem dependent \cite{li2014efficient}.

Alternative ways to scale up variational inference in conjugate exponential models involve the use of distributed computing clusters. For example, it can be assumed that the data set is stored in a distributed way among different machines  \cite{masegosa2017scaling}. Then the problem of computing the ELBO's gradient given in Equation~\eqref{eq:gradELBO} is scaled up by distributing the computation of the gradient  $\nabla^{nat}_{\bm \phi_i}\lb (\bmlambda,\bmphi)$ so that each machine computes this term for those samples that are locally stored. Finally, all the terms are sent to a master node which aggregates them and compute the gradient $\nabla^{nat}_{\bm \lambda} \lb (\bmlambda,\bmphi)$ (see Equation~\eqref{eq:gradELBO}).

\begin{exmp}\label{example:PCA:SVI}
In Example  \ref{example:PCA:VI} we detailed the variational updating equations for the Probabilistic PCA model introduced in Example \ref{example:PCA}. 
In order to update $\bmmu_{\bm \beta}^\star$  we need to iterate over the whole data set. 
Furthermore, the number of local variational parameters $\bmmu_{z_i}^\star$ and $\Sigma_{z_i}^\star$ is equal to the number of data points. 
Therefore, if $N$ is very large, the computation of these variational updating equations becomes infeasible. 

Following the methodology  presented in this section, we can obtain a new set of variational updating equations, 
\begin{eqnarray*}
\label{pca:vi-ii}
\bmSigma_{\beta,t+1} &=& \left[\left(1-\rho_t\right) \bmSigma^{-1}_{\beta,t} +\rho_t\left(\frac{N}{M}\sum_{m=1}^M \E_{t, i_m}\left[\bmZ_{i_m}\bmZ_{i_m}\trans\right] +  \sigma_x^2 \bmA\right)\right]^{-1} , \\
\bmmu_{\beta,t+1} &=&  \left(1-\rho_t\right)\bmmu_{\beta,t+1} + \rho_t\left(\frac{N}{M}\sum_{m=1}^M \bmx_{i_m} \E_{t, i_m}[\bmZ_{i_m}]\right)\trans\bmSigma_{\beta,t+1} ,
\end{eqnarray*}  
\noindent where $\{i_1\ld i_M\}$ are the indexes of the mini-batch, and $\E_{t, i_m}[\cdot]$ denotes expectations when $\bmZ_{i_m}$ follows a Normal distribution with parameters
\begin{eqnarray*}
\bmSigma_{t, z_{i_m}} &=& \left(\bmI + \sigma_x^{-2}\bmmu_{\beta, t}\trans\bmmu_{\beta, t} \right)^{-1}, \\
\bmmu_{t, z_{i_m}} &=& \bmSigma_{t, z_{i_m}} \bmmu_{\beta, t}\trans \bmx_{i_m} / \sigma_x^2;
\end{eqnarray*}  
\noindent
confer also Example \ref{example:PCA:VI}.
Using this set-up , we do not need to go thorough the full data set to get an update of the global variational parameters. 
\end{exmp}

\subsection{Variational Message Passing}\label{sec:VMP}

So far we have treated the set of variables $\bmx$, $\bmz$ and $\bmbeta$ as undividable blocks of variables without internal structure. 
However, as we are dealing with flexible probabilistic graphical models,  these sets of variables can often encode conditional independencies that can be further exploited when using VI. 
\textit{Variational message passing} (VMP) \cite{WinnBishop05} is a VI scheme which readily exploits such conditional independencies when performing approximate inference. 
Now, $\bmZ_i$ and $\bmX_i$, the set of latent and observable variables associated to the $i$-th data sample, are separated into 
 individual variables $\bmZ_i=\{Z_{i,1}\ld Z_{i,K}\}$, and similarly for $\bmX_i$. 
 Additionally, $\bmbeta$ is regarded as a set of $J$ separate random variables $\bmbeta=\{\beta_1,\ldots,\beta_J\}$. 
 Now, under the mean field assumption, the variational distribution is expressed as 
\[
q(\bmbeta,\bmz|\bmlambda_i,\bmphi) = \prod_{j=1}^J q(\beta_j|\bmlambda_j) \prod_{i=1}^N\prod_{k=1}^K  q(z_{i,k}|\bmphi_{i,k}) . 
\] 

Using the VMP scheme, the gradients wrt.\ the variational parameters can be computed using a message-passing algorithm which exploits the conditional independencies between the variables in $\bmX_i$, $\bmZ_i$ and $\bmbeta$. 
The flow of messages is similar to the one employed by loopy belief propagation \cite{Pearl88}. 
The messages are expected sufficient statistics of the variables involved, and since the model is in the conjugate exponential family, both the messages and the update rules can be expressed analytically, 
leading to parameter updates akin to Equation \eqref{eq:CoordinateAscent}; cf.\ \cite{WinnBishop05} for details.

\vekk{

\comment{HL: I think this subsection is borderline unreadable. I still find the original paper hard to read, even if I know the details. 
This text is even  more difficult, partly because it is too compressed, and it also opens up for discussions that are not made to their conclusions. 
For instance, it is not clear what the message is, what messages are to be sent and how they are received, how co-parents are to interact with messages from children, etc. 
My suggestion is to remove the text from here on close it down by saying something like ``details in Winn and Bishop''. Alternatively, delete the full subsection.}

For instance is the gradient of the ELBO with respect to $\bmphi_{i,k}$ computed as a sum of messages coming only from those variables belonging to the Markov blanket of $Z_{i,k}$, 
\[ 
\nabla_{\phi_{i,k}}\lb  = m_{\pa{z_{i,k}} \rightarrow z_{i,k}} + \sum_{y \in ch(z_{i,k})} m_{ y \rightarrow z_{i,k}} (co_{z_{i,k}}(y)) - \bmphi_{i,k}, 
\]
\noindent 
where $\pa{z_{i,k}}$ and $ch(z_{i,k})$ denotes the set of parents and children of $z_{i,k}$, respectively (i.e. those random variables with a link pointing to $z_{i,k}$ or with an incoming link departing from to $z_{i,k}$, respectively). In addition, $m_{ y \rightarrow z_{i,k}} (co_{z_{i,k}}(y))$ denotes that the message that $y$ sends to $z_{i,k}$ depends of the set of co-parents of $y$ wrt.\ $z_{i,k}$, denoted by $co_{z_{i,k}}(y)$ (i.e. the set of of parents of $y$ excluding $z_{i,k}$). All these messages can be computed as expected sufficient statistics of the variables involved, where the expectation is taken wrt.\ to the corresponding variational distribution. See \cite{WinnBishop05} for more details. The same idea applies when computing the gradient of $\lb$ wrt.\ $\beta_j$. In fact, under VMP there is no distinction between global and local variational parameters, all are treated exactly in the same way. 


}

%% file: dnn.tex
\section{Deep Neural Networks and Computational Graphs}\label{sec:deepframeworks}

\subsection{Deep Neural Networks}

An artificial neural network (ANN) \citep{hopfield1988artificial} can be seen as a deterministic non-linear function $f(\cdot: \bmW)$ parametrized by a matrix $\bmW$. An ANN with $L$ hidden layers defines  a mapping from a given input $\bmx$ to a given output $\bmy$. This mapping is built by the recursive application of a sequence of (non-)linear transformations, 
\begin{eqnarray}
\label{eq:ANNs}
\bmh_0 &=& \bmr_0(\bmW_0^T\bmx) , \nonumber\\
&\ldots &\nonumber\\
\bmh_{l} &=& \bmr_l(\bmW_{l-1}^T\bmh_{l-1}) , \nonumber\\
&\ldots &\nonumber\\
\bmy &=& \bmr_L(\bmW_L^T\bmh_{L}) , 
\end{eqnarray} 
\noindent where $\bmr_l(\cdot)$ defines the (non-linear) activation function at the $l$-th layer; standard
activation functions include the \textit{soft-max} function and the \textit{relu} function
\citep{goodfellow2016deep}. $\bmW_l$ are the parameters defining the linear transformation at the $l$-th
layer, where the dimensionality of the target layer is defined by the size of $\bmW_l$. Deep neural networks (DNNs) is a renaming of classic ANNs, with the key difference that DNNs usually
have a higher number of hidden layers compared to what classical ANNs used to have. 


Learning a DNN from a given data set of input-output pairs $(\bmx,\bmy)$ reduces to solving the optimization problem 
\begin{equation}
\label{eq:dnnlearning}
\bmW^\star = \arg\min_\bmW \sum_{i=1}^N \ell (y_i,f(\bmx_i;\bmW)),
\end{equation}
\noindent where $\ell (\bmy_i,\hat{\bmy}_i)$ is a loss function which defines the quality of the model, i.e,
how well the output $\hat{\bmy}_i = f(\bmx_i;\bmW)$ returned by the DNN model matches the real output
$\bmy_i$. This continuous optimization problem is usually solved by applying a variant of the stochastic
gradient descent method, which involves the computation of the gradient of the loss function with respect to
the parameters of the ANN,  $\nabla_\bmW \ell (y_i,f(\bmx_i;\bmW))$. The algorithm for computing this gradient
in an ANN is known as the \textit{back-propagation} algorithm, which is based on the recursive application of
the chain-rule of derivatives and typically implemented based in the computational graph of the
ANN.  A detailed and modern introduction to this field is provided in \cite{goodfellow2016deep}.

\subsection{Computational Graphs}
\begin{figure}[htb!]
\centering

\begin{tikzpicture}

\tikzstyle{stoc}=[circle, minimum size = 10mm, thick, draw =black!80, node distance = 16mm]
\tikzstyle{deter}=[rectangle, minimum size = 10mm, thick, draw =black!80, node distance = 16mm]
\tikzstyle{input}=[minimum size = 5mm, node distance = 3mm]

\tikzstyle{connect}=[-latex, thick]

\node[input] (w0) at (2,2)  {$w$};
\node[input] (w1) at (5,2)  {$10$};
\node[input] (x) at (0,0)  {$3$};
\node[deter] (h0) at (2,0)  {$*$};
\node[deter] (h1) at (5,0)  {$+$};
\node[input] (f) at (7,0)  {$f$};

\path (w0) [connect]  edge (h0);
\path (w1) [connect]  edge (h1);
\path (x) [connect]  edge (h0);
\path (h0) [connect]  edge (h1);
\path (h1) [connect]  edge (f);

\end{tikzpicture}

\caption{\label{fig:simplecomputationalgraph} Example of a simple Computational Graph. Squared nodes denote operations, and the rest are input nodes. This computational graph encodes the operation $f = 3\cdot w + 10$, where $w$ is a variable wrt.\ which we can differentiate.}

\end{figure}
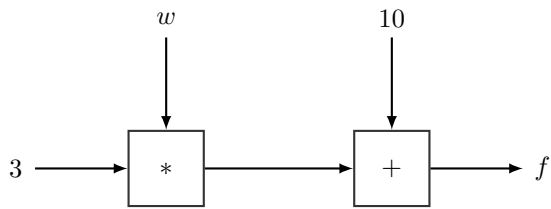

Computational graphs have been extremely useful when developing algorithms and software packages for neural
networks and other models in machine learning
\citep{chen2015mxnet,tensorflow2015-whitepaper,paszke2017automatic}. The main idea of a computational graph is
to express a (deterministic) function, as is the case of a neural network, as an acyclic directed graph
defining a sequence of computational operations. A computational graph is composed of input and output nodes
as well as operation nodes. The data and the parameters of the model serve as input nodes, whereas the
operation nodes (represented as squares in the subsequent diagrams) correspond to the primitive operations of
the network and also define the output of the network. The directed edges in the graph specify the
inputs of each node. Input nodes are usually defined over tensors ($n$-dimensional arrays) and operations are
thus similarly defined over tensors, thereby also enabling the computational graph to, e.g., process batches of data. Figure
\ref{fig:simplecomputationalgraph} shows a simple example of a computational graph.

With computational graphs, simple/primitive functions can be combined to form complex operations, and the vast majority of current neural networks can be defined using computational
graphs. But the key strength of computational graphs is that they allow for automatic differentiation
\citep{griewank1989automatic}. As shown in the previous section (see Equation \eqref{eq:dnnlearning}), most
neural network learning algorithms translate to a continuous optimization problem of a differentiable loss
function often solved by a gradient descent algorithm. 
Automatic differentiation is a technique for automatically computing all the partial derivatives of the
function encoded by a computational graph: once the graph has been defined using underlying primitive
operations, derivatives are automatically calculated based on the ``local'' derivatives of these operations
and the recursive application of the chain rule of derivatives, incurring only a small computational overhead. Before the
use of computational graphs in deep learning, these derivatives had to be computed manually, giving rise to a
slow and error-prone development process. 

\begin{figure}[htb!]

\centering

\begin{tikzpicture}

\tikzstyle{stoc}=[circle, minimum size = 10mm, thick, draw =black!80, node distance = 16mm]
\tikzstyle{deter}=[rectangle, minimum size = 10mm, thick, draw =black!80, node distance = 16mm]
\tikzstyle{input}=[minimum size = 5mm, node distance = 3mm]

\tikzstyle{connect}=[-latex, thick]

\node[input] (w0) at (2,2)  {$\bmW_{0}$};
\node[input] (w1) at (5,2)  {$\bmW_{1}$};
\node[input] (w2) at (8,2)  {$\bmW_{2}$};
\node[input] (y) at (11,2)  {$\bmy$};

\node[input] (x) at (0,0)  {$\bmx$};
\node[deter] (h0) at (2,0)  {\footnotesize{$\bmh_{0} = relu(\bmW_{0}^{T}\bmx )$}};
\node[deter] (h1) at (5,0)  {\footnotesize{$\bmh_{1} = relu(\bmW_{2}^{T}\bmh_{0} )$}};
\node[deter] (est_y) at (8,0)  {\footnotesize{$\hat{\bmy} = relu(\bmW_{2}^{T}\bmh_{1} )$}};
\node[deter] (l) at (11,0)  {\footnotesize{$\ell = ||\hat{\bmy} - \bmy||^{2}$}};

\path (w0) [connect]  edge (h0);
\path (w1) [connect]  edge (h1);
\path (w2) [connect]  edge (est_y);
\path (y) [connect]  edge (l);

\path (x) [connect]  edge (h0);
\path (h0) [connect]  edge (h1);
\path (h1) [connect]  edge (est_y);
\path (est_y) [connect]  edge (l);

\end{tikzpicture}

\caption{\label{fig:computationalgraph} Example of a simple Computational Graph encoding a neural network with two hidden layers and and the squared loss function. Note that each operation node encapsulates a part of the CG encoding the associated operations, we do not expand the whole CG for the sake of simplicity.}

\end{figure}
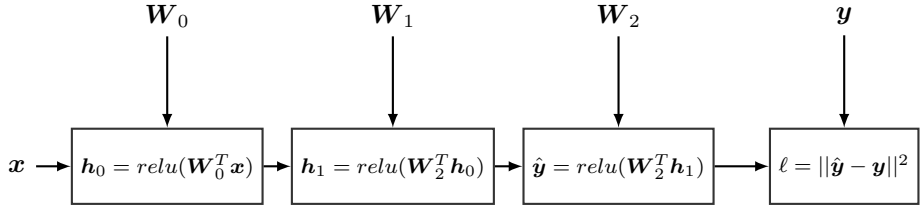

\begin{exmp}
Figure~\ref{fig:computationalgraph} provides an example of a computational graph encoding a neural network
with $\bmx$ as input, $\hat{\bmy}$ as output, and two hidden layers. This computational graph also encodes the
loss function $\ell(\bmy,\hat{\bmy})$. As computational graphs can be defined over tensors, the above
computational graph can encode the forward (and backward) pass of the neural network for a whole data batch
$\bmx$, and thereby also provide the loss (and the gradient) for this set of data samples.  Algorithm
\ref{alg:nn} shows the pseudo-code description for defining and learning this neural network using standard
gradient descent. 

\begin{algorithm}[htb!]
\caption{Pseudo-code of the definition and learning of a simple neural network.}\label{alg:nn}
\begin{algorithmic}
\INPUT $\bmx$, $\bmy$ 
the labels.
\STATE \# Define the computational graph encoding the ANN and the loss function
\STATE $\bmW_0, \bmW_1, \bmW_2  = \emph{Parameters}()$
\STATE $\bmh_0 = relu(\bmW_0^T\bmx )$ 
\STATE $\bmh_1 = relu(\bmW_1^T\bmh_0)$ 
\STATE $\hat{\bmy} = relu(\bmW_2^T\bmh_1)$ 
\STATE $\ell$ = $||\hat{\bmy} - \bmy||^2$ 
\STATE \# Follow the gradients until convergence.
\STATE $\bmW = (\bmW_0,\bmW_1,\bmW_2)$
\REPEAT
\STATE $\bmW = \bmW - \rho \nabla_{\bmW}\ell$  
\UNTIL{convergence}
\end{algorithmic}
\end{algorithm}
%

\end{exmp}

%% file: deeplvm.tex
\section{Probabilistic Models with Deep Neural networks}\label{sec:modelsDNNs}
\subsection{Deep Latent Variable Models}\label{sec:deepLVMs}
LVMs have usually been restricted to the conjugate exponential family because, in this case,  inference is
feasible (and scalable) as we showed in Section \ref{sec:models}. But recent advances in VI (which will be
reviewed in Section \ref{sec:variational}) have enabled LVMs to be extended with
DNNs. Variational Auto-encoders (VAE) \citep{kingma2013auto,doersch2016tutorial} are probably the most
influential models combining LVMs and DNNs. VAEs extend the classical technique of
PCA
for data representation in lower-dimensional spaces. More
precisely, the probabilistic version of the PCA model
\citep{tipping1999probabilistic} is extended in \cite{kingma2013auto}, where the relationship between the low-dimensional representation and the
observed data is governed by a DNN, i.e. a highly non-linear function, as opposed to the standard linear
transformation in the basic version of the PCA model. These models are able to capture more compact
low-dimensional representations, especially in cases where data is high-dimensional but ``lives'' in a low-dimensional manifold
  \citep{pless2009survey}. This is, e.g., the case for image data \citep{kingma2013auto,kulkarni2015deep,gregor2015draw,sohn2015learning,pu2016variational}, text data \citep{semeniuta2017hybrid}, audio data \citep{hsu2017learning}, chemical molecules \citep{gomez2018automatic}, to name some representative applications of this technique. We note that, in this section and in the following ones, we will use VAEs as a running example illustrating how DNNs can be used in probabilistic modeling. 

VAEs have also given rise to a plethora of extensions of classic LVMs to their \emph{deep} counterpart. For
instance,  different examples of this approach are given in~\cite{johnson2016composing}, along with extensions of
instance, provides different examples of this approach are given in~\cite{johnson2016composing}, along with extensions of
Gaussian mixture models, latent linear dynamical systems and latent switching linear dynamical systems with
the non-linear relationships modelled by DNNs. Hidden semi-Markov models are extended with 
 recurrent neural networks in  \cite{linderman2016recurrent}.  Extensions of popular LDA models \citep{blei2003latent} for uncovering topics in text data can be found in \cite{zhou2015poisson} and \cite{card2017neural}. Many other works are following the same trend \citep{chung2015recurrent,jiang2016variational,xie2016unsupervised,louizos2017causal}.

\begin{figure}[htb!]
	\resizebox{\linewidth}{!}{
		\begin{tabular}{cc}
			\includegraphics{pca_mnist.pdf}
			&
			\includegraphics{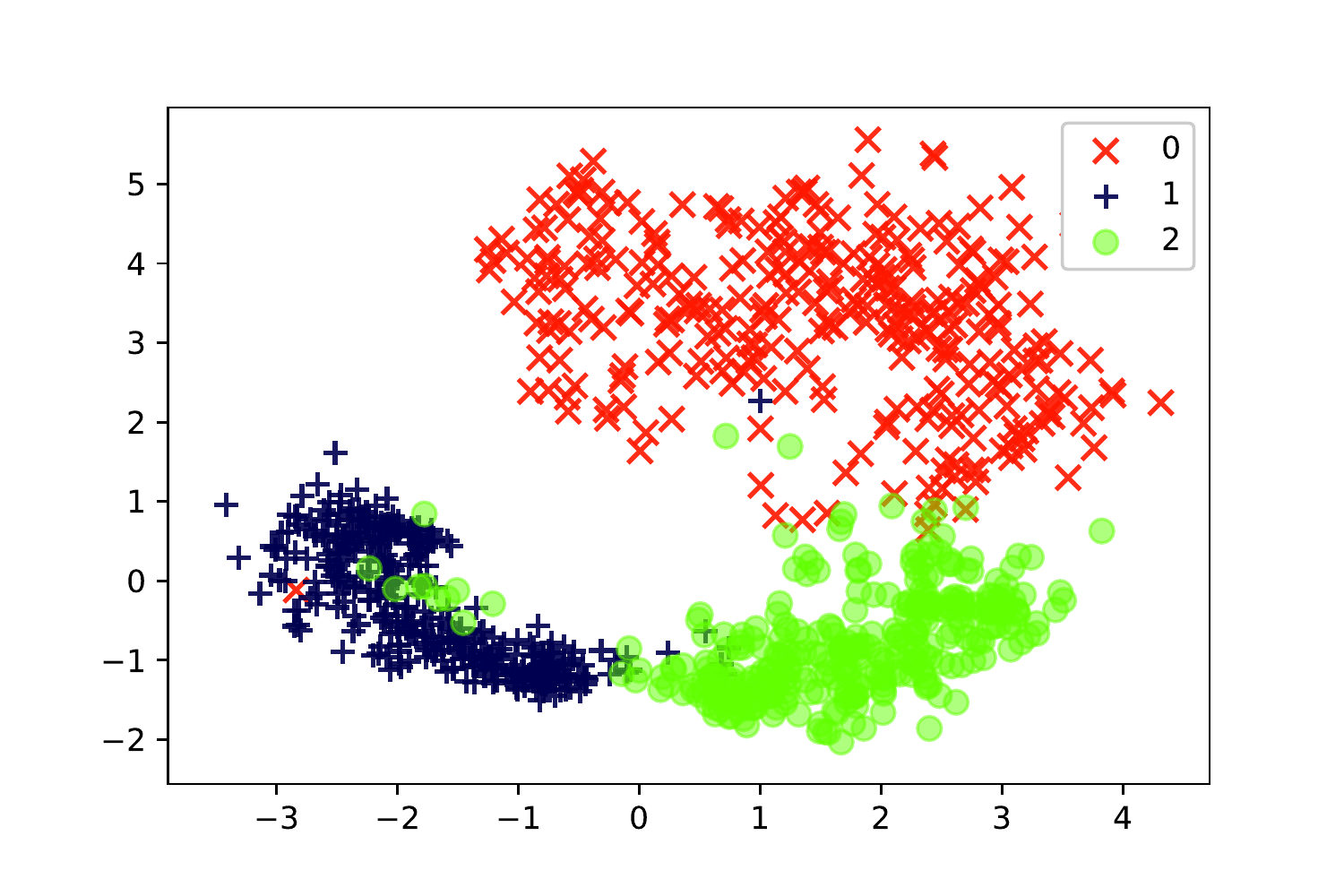}
		\end{tabular}
	}
	\caption{\label{fig:PCANonLineal:Application} 2-dimensional latent representations of the the MNIST
          dataset resulting of applying:  \textbf{(Left)} a standard probabilistic PCA (reproduced from
          Figure~\ref{fig:PCA:application} to ease comparison), and \textbf{(Right)} a non-linear probabilistic PCA with a ANN containing a hidden layer of size 100  with a \textit{relu} activation function.}
\end{figure}

\begin{exmp}\label{example:vae}
VAEs are widely adopted LVMs containing DNNs \citep{kingma2013auto}. Algorithm
\ref{alg:vae} provides a simplified pseudo-code description of the \textit{generative part} of a VAE model. It
can also be seen as a non-linear probabilistic PCA model, where the non-linearity is included in the form of
an artificial neural network. 


%

This model is quite similar to the PCA model presented in Example~\ref{example:PCA}. The main difference comes
from the conditional distribution of $\bmX$. In the PCA model, the mean of the normal distribution of $\bmX$
linearly depends  on $\bmZ$ through $\bmbeta$. In the VAE model, the mean depends on $\bmZ$ through a DNN
parametrized by $\bmbeta$. This DNN is also known as the \textit{decoder network} of the VAE \citep{kingma2013auto}. 

Note that some formulations of this model also include another DNN component, which connects $\bmZ$ with the variance
$\sigma^2$ of the Normal distribution of $\bmX$; for the sake of simplicity, we have not included this
extension in the example. 


Figure \ref{fig:PCANonLineal:Application} experimentally illustrates the advantage of using a non-linear PCA
model over the classic PCA model. As can be seen, the non-linear version separates more clearly the three
digits than the linear model did.
We shall return to this example in Section~\ref{sec:elbo-optim-with}, where we will
introduce the so-called \emph{encoder network} used for inference.

\end{exmp}

\begin{algorithm}[htb!]
\caption{\label{alg:nonlinearPCA}Pseudo-code of the generative model of a Variational Auto-encoder (or non-linear probabilistic PCA)}\label{alg:vae}
\begin{algorithmic}
\STATE \# Define the global parameters
\STATE $\bmalpha_0, \bmbeta_0, \bmalpha_1, \bmbeta_1 \sim {\cal N} (\bm0,\bmI)$
\FOR{$i=1\ld N$}
\STATE \# Define the local latent variables 
\STATE $\bmZ_i \sim {\cal N} (\bm0,\bmI)$ 
\STATE \# Define the ANN with a single hidden layer $\bmh_i$
\STATE $\bmh_i = relu(\bmbeta_0^T\bmz_i  + \bmalpha_0)$ 
\STATE $\bmmu_{i} = \bmbeta_1^T\bmh_i + \bmalpha_1$
\STATE \# Define the observed variables 
\STATE $\bmX_i \sim {\cal N} (\bmmu_{i}, \sigma^2\bmI)$ 
\ENDFOR
\end{algorithmic}
\end{algorithm}

LVMs with DNNs can also be found in the literature under the name of deep generative models
\citep{hinton2009deep,hinton2012practical,goodfellow2014generative,salakhutdinov2015learning}. 
They 
generate data samples using probabilistic constructs that include DNNs. 
This new capacity has also 
resulted in substantial 
impact within the deep learning community because it has opened up for the
possibility of dealing with unsupervised learning problems, e.g., in the form of generative adversarial nets~\cite{goodfellow2014generative}. This should be seen in contrast to the \textit{classic deep learning
  methods}, which are mainly focused on supervised learning settings. In any case, this active area of research is out of the scope of this paper and contains many alternative models, which do not fall within the category of the models explored in this paper.



\subsection{Stochastic Computational Graphs}

One of the main reasons for the wide adoption of deep learning has been the availability of (open-source) software tools containing robust and well-tested implementations of the main building blocks for defining and learning DNNs \citep{tensorflow2015-whitepaper,paszke2017automatic}. Recently, a new wave of software tools have appeared, building on top of these deep learning
frameworks in order to accommodate modern probabilistic models containing DNNs
\citep{tran2016edward,cabanasInferPy,cozar2019inferpy,tran2018simple,bingham2018pyro}. These software tools usually fall under
the umbrella term \textit{probabilistic programming languages} (PPLs)
\citep{gordon2014probabilistic,ghahramani2015probabilistic}, and  support methods for learning and reasoning about complex
probabilistic models. Although PPLs have been present in the field of machine learning for many years,
traditional PPLs have mainly focused on defining languages for expressing (more restricted types of)
probabilistic models \citep{koller2009probabilistic} with only little focus on issues such as scalability. The
advent of deep learning and the introduction of probabilistic models containing DNNs has motivated the
development of a new family of PPLs offering support for flexible and complex models as well as scalable inference. Well-known examples include Edward2 \citep{tran2016edward,tran2018simple}, developed by Google and built on top of TensorFlow \citep{tensorflow2015-whitepaper}, Pyro \citep{bingham2018pyro} developed by Uber and built on top of Pytorch \citep{paszke2017automatic}, and PyMC3 \citep{salvatier2016probabilistic} which is built on top of Theano \citep{bergstra2010theano}.


%
%
%
%
%
%
%
%
%
%
%
%
%
%

The key data structure in these new PPLs are the so-called \textit{stochastic computational graphs} (SCGs)
\citep{schulman2015gradient}. SCGs extend standard computational graphs with stochastic nodes
(represented as circles in the subsequent diagrams). The probability distributions associated with stochastic
nodes are defined conditionally on their
parents and enable the specification of complex functions involving expectations over random
variables. Figure~\ref{fig:EvaluatingStochasticCG} \textbf{(Left)} shows an example of a simple SCG involving
an expectation over a random variable $\bmZ$. Modern PPLs support a wide and diverse range of probability distributions for defining SCGs \citep{dillon2017tensorflow}. These probability distributions are defined over tensor objects to seamlessly accommodate the underlying CGs, which define operations over tensor objects too. 

We note that SCGs are not directly implemented within PPLs, because computing the exact expected
value of a complex function is typically infeasible. However, they are indirectly included through the use
of a standard
computational graph engine: Each stochastic node, $\bmZ$, is associated with a tensor,
$\bmz^\star$, which represents a (set of) sample(s) from the distribution associated with $\bmz$, and the
generated samples can thus be fed to the underlying computational graph through the tensor
$\bmz^\star$. Hence, SCGs can be \textit{simulated} by sampling from the stochastic nodes and processing these
samples by a standard CG. Figure \ref{fig:EvaluatingStochasticCG} illustrates how SCG can be simulated using
standard CGs. Note that 
CGs are designed to operate efficiently with tensors (current toolboxes like TensorFlow exploit high-performance computing hardware such as GPUs or TPUs 
\citep{tensorflow2015-whitepaper}), and it is therefore much more efficient to run the CG once over a collection of
samples, rather than running the CG multiple times over a single 
sample.

\begin{figure}[htb!]

\centering

\begin{tikzpicture}

\tikzstyle{stoc}=[circle, minimum size = 10mm, thick, draw =black!80, node distance = 16mm]
\tikzstyle{deter}=[rectangle, minimum size = 10mm, thick, draw =black!80, node distance = 16mm]
\tikzstyle{input}=[minimum size = 5mm, node distance = 3mm]

\tikzstyle{connect}=[-latex, thick]

\node[input] (h2) at (7,3)  {$\hat{h} = \frac{1}{k}\sum_{i=1}^{k}(z_{i}^{\star}-5)^2, \quad Z_i^\star{\sim}\calN(\mu,1)$};

\node[input] (Ez) at (1,3)  {$h = \E_{Z\sim \calN(\mu,1)}[(Z-5)^2]$};

\node[deter] (h3) at (1,5.5)  {$(Z-5)^2$};
\node[deter] (h5) at (7.5,5.5)  {$(z^\star-5)^2$};
\node[deter] (h5avg) at (5.5,5.5)  {avg};

\node[input] (h3_) at (1,4)  {$h$};
\node[input] (h5_) at (5.5,4)  {$\hat{h}$};

\node[stoc] (z3) at (0.5,7)  {$Z$};
\node[input] (number_1) at (1.5,7)  {5};

\node[stoc] (z6) at (5.5,7)  {$Z$};
\node[deter] (z7) at (7,7)  {$z^{\star}$};
\node[input] (number_3) at (8,7)  {5};

\node[input] (lambda1) at (0.5,8.5)  {$\mu$};
\node[input] (lambda3) at (5.5,8.5)  {$\mu$};

\path (h3) [connect]  edge (h3_);
\path (h5) [connect]  edge (h5avg);
\path (h5avg) [connect]  edge (h5_);

\path (z3) [connect]  edge (h3);
\path (number_1) [connect]  edge (h3);


\path (z6) [connect]  edge (z7);
\path (z7) [connect]  edge (h5);
\path (number_3) [connect]  edge (h5);

\path (lambda1) [connect]  edge (z3);
\path (lambda3) [connect]  edge (z6);

\end{tikzpicture}

\caption{\label{fig:EvaluatingStochasticCG} \textbf{ (Left)} A stochastic computational graph encoding the function $h=\E_{Z}[(Z - 5)^2]$, where $Z\sim 
N(\mu,1)$.  \textbf{(Right)}  Computational graph processing $k$ samples from $Z$ and producing $\hat{h}$, an estimate of  $\E_Z[(Z - 5)^2]$. }
\end{figure}

\begin{figure}
\begin{center}

\scalebox{1.1}{
\begin{tikzpicture}
  \node[latent]  (beta)   {$\boldsymbol{\beta}$}; %
  
  
  \node[obs, below right=1 of beta]          (X)   {$\bmX_{i}$}; %
  \edge{beta}{X};
 
  \node[latent, below left= 1 of beta]          (Z)   {$\bmZ_{i}$}; %
  \edge{Z}{X};
  \edge{beta}{Z};
  
  \plate {Observations} { %
    (X)(Z)
  } {\tiny $i={1,\ldots, N}$} ;


\end{tikzpicture}
}


\vspace*{0.5cm}
\begin{tikzpicture}[thick, scale=0.8, every node/.style={scale=0.8}]

\tikzstyle{stoc}=[circle, minimum size = 10mm, thick, draw =black!80, node distance = 16mm]
\tikzstyle{deter}=[rectangle, minimum size = 10mm, thick, draw =black!80, node distance = 16mm]
\tikzstyle{input}=[minimum size = 5mm, node distance = 3mm]

\tikzstyle{connect}=[-latex, thick]

\node[stoc] (z) at (0,0)  {$\bmz$};
\node[deter] (h0) at (2,0)  {$\bmh_{0}$};
\node[input] (ellipsis) at (4, 0)  {$\cdots$};
\node[deter] (hL_1) at (6,0)  {$\bmh_{L-1}$};
\node[deter] (hL) at (8,0)  {$\bmh_{L}$};
\node[deter] (x) at (10,0)  {$\bmeta_x$};
\node[stoc] (xtrue) at (12,0)  {$\bmx$};

\node[stoc] (B1) at (0,2)  {$\bmbeta_{z}$};
\node[stoc] (B0) at (2,2)  {$\bmbeta_{0}$};
\node[stoc] (BL_1) at (6,2)  {$\bmbeta_{L-1}$};
\node[stoc] (BL) at (8,2)  {$\bmbeta_{L}$};


\path (z) [connect]  edge (h0);
\path (h0) [connect]  edge (ellipsis);
\path (ellipsis) [connect]  edge (hL_1);
\path (hL_1) [connect]  edge (hL);
\path (hL) [connect]  edge (x);
\path (x) [connect]  edge (xtrue);

\path (B1) [connect]  edge (z);
\path (B0) [connect]  edge (h0);
\path (BL_1) [connect]  edge (hL_1);
\path (BL) [connect]  edge (hL);


\end{tikzpicture}

\end{center}
\caption{\label{fig:DNNModel}The top part depicts a probabilistic graphical model using plate notation
  \citep{koller2009probabilistic}. The lower part depicts an abstract representation of a stochastic
  computational graph encoding the model, where the relation between $\bmz$ and $\bmx$ is defined by a DNN
  with $L+1$ layers. See Section \ref{sec:modelsDNNs} for details.
}
\end{figure}
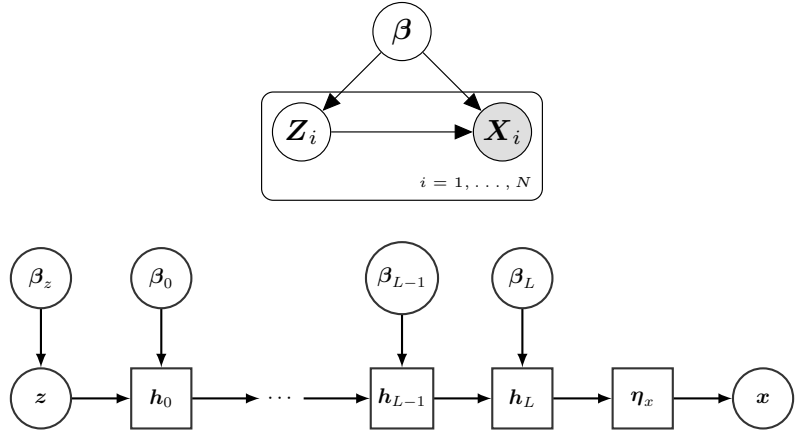

In this way, SCGs can be used to define and support inference and learning of general probabilistic models, including the ones
referenced in Section~\ref{sec:deepLVMs}. More generally, all the concepts reviewed in this paper apply to any
probabilistic model that can be defined by means of an SCG or which can be compiled into an equivalent SCG
representation. For instance, the following model specification (illustrated by the top part in
Figure~\ref{fig:DNNModel}) relating $\bmZ$ with the natural parameters
$\bmeta_x$ of $\bmx$ can be equivalently represented by the SCG illustrated in the lower part of
Figure~\ref{fig:DNNModel}.

\begin{eqnarray}\label{eq:dnnexponentialform}
\ln p(\bmbeta) &=& \ln h(\bmbeta) + \bmalpha^T t(\bmbeta) - a_g(\bmalpha), \nonumber\\
\ln p(\bmz_i|\bmbeta) &=& \ln h(\bmz_i)  + \eta_z(\bmbeta)^T t(\bmz_i) - a_z(\eta_z(\bmbeta)), \nonumber\\
\bmh_0 &=& \bmr_0(\bmz_i^T\bmbeta_0), \nonumber\\
&\ldots &\nonumber\\
\bmh_{l} &=& \bmr_l(\bmh_{l-1}^T\bmbeta_{l-1}), \nonumber\\
&\ldots &\nonumber\\
\bmh_L &=& \bmr_L(\bmh_{L}^T\bmbeta_L), \nonumber\\
\ln p(\bmx_i|\bmz_i,\bmbeta) &=& \ln h(\bmx_i)  + \eta_x(\bmh_L)^T t(\bmx_i) - a_x(\eta_x(\bmh_L)).
\end{eqnarray}

%

From this example, we again see the main difference with respect to standard LVMs (see Section \ref{sec:LVM})
is the conditional distribution of the observations $\bmx_i$ given the local hidden variables $\bmZ_i$ and the
global parameters $\bmbeta$, which is here governed by a DNN parameterized by $\bmbeta$. 



%% file: variational.tex
\section{Variational Inference with Deep Neural Networks}\label{sec:variational}

Similarly to standard probabilistic models, performing variational inference in deep latent variable models (as described in the
previous section) also reduces to maximizing the ELBO function ${\cal L}(\bmlambda,\bmphi)$ given in Section
\ref{sec:VI} (Equation~\eqref{eq:elbo}); recall that this is equivalent to minimizing the KL divergence
between the variational posterior $q(\bmbeta,\bmz\given\bmlambda,\bmphi)$ and the target distribution
$p(\bmbeta,\bmz\given\bmx)$. However, as was also noted in the previous section, when the probabilistic model
contains complex constructs like DNNs, it falls outside the conjugate exponential family and the traditional
VI methods, tailored to this specific family form, can therefore not be applied.  


In terms of the variational distribution, we will for the deep latent variable models considered in this
section still assume the same factorization scheme defined in
Equation~\eqref{eq:meanfieldQ}. However, as we will see below, we need not adopt the conjugate models' strong
restrictions on
the variational approximation family (see Equation~\eqref{eq:qcompleteconditionals}). Instead, the only (and much weaker) restriction that we will impose is that $i$)
the log probability of the variational distribuiton, $\ln q(\bmbeta,\bmz\given\bmlambda,\bmphi)$, can be
represented by a computational graph (and, as a consequence, that it is differentiable wrt.\ $\bmlambda$ and
$\bmphi$) and $ii$) that we can sample from the variational distribution $q(\bmbeta,\bmz\given\bmlambda,\bmphi)$.
Depending on the specific method being applied, additional requirements may be introduced. The main methods currently
available in the literature are introduced in the rest of this section.  


\subsection{Black Box Variational Inference}
\label{sec:blackbox}

\begin{figure}[htb!]

\centering

\begin{tikzpicture}

\tikzstyle{stoc}=[circle, minimum size = 10mm, thick, draw =black!80, node distance = 16mm]
\tikzstyle{deter}=[rectangle, minimum size = 10mm, thick, draw =black!80, node distance = 16mm]
\tikzstyle{input}=[minimum size = 5mm, node distance = 3mm]

\tikzstyle{connect}=[-latex, thick]

\node[stoc] (r) at (0,0)  {$\bmr$};
\node[deter] (g) at (2,0)  {$g$};

\node[input] (v) at (0,2)  {$\bmnu$};
\node[input] (x) at (2,2)  {$\bmx$};

\path (r) [connect]  edge (g);

\path (v) [connect]  edge (r);
\path (v) [connect]  edge (g);

\path (x) [connect]  edge (g);

\end{tikzpicture}
\vspace{-5pt}
\caption{\label{fig:ELBOSCG} SCG representing the ELBO function ${\cal L}(\bmnu)$. $\bmr$ is distributed according to the variational distribution, $\bmr \sim q(\bmr\given\bmnu)$.  }

\end{figure}
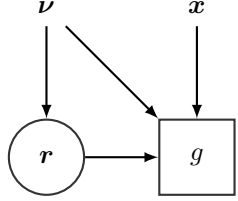

For the sake of presentation, we reparameterize the ELBO function with $\bmr=(\bmbeta,\bmz)$ and $\bmnu =
(\bmlambda,\bmphi)$ and define $g(\bmr,\bmnu) = \ln p(\bmx,\bmr) - \ln q(\bmr\given\bmnu)$. With this
notation the ELBO function ${\cal L}$ of Equation~\eqref{eq:elbo} can then be expressed as
\begin{equation}
  \label{eq:elbo_dnn}
{\cal L}(\bmnu) = \E_{\bmR}[g(\bmr,\bmnu)] = \int q(\bmr\given\bmnu) g(\bmr,\bmnu) d\bmr ,
\end{equation}
from which we see that the ELBO function can easily be represented by an SCG as shown in Figure
\ref{fig:ELBOSCG}. If the SCG in Figure \ref{fig:ELBOSCG} did not include stochastic nodes (thus corresponding
to a standard CG), we could in principle perform variational inference (maximizing ${\cal L}(\bmnu)$ wrt.\
$\bmnu$) by simply relying on automatic differentiation and a variation of gradient ascent. However,
optimizing over SCGs is much more challenging because automatic differentiation does not readily apply. The
problem is that the variational
parameters $\bmnu$ (wrt.\ which we should differentiate) also affects the expectation inherent in the ELBO
function, see Equation~(\ref{eq:elbo_dnn}):
\begin{equation}\label{eq:gradELBODNNs}
\nabla_{\bm \nu} {\cal L}=  \nabla_{\bm \nu} \E_{\bmR}[g(\bmr,\bmnu)].
\end{equation}
In the case of conjugate models, we can take advantage of their properties and derive closed-form solutions
for this problem, as detailed in Section \ref{sec:VI}. In general, though, there are no closed-form solutions
for computing gradients in non-conjugate models; a simple concrete example is the  Bayesian logistic regression model
\citep[Page 756]{murphy2012machine}. 

In this section, we provide two generic solutions for computing the gradient of the ELBO function for
probabilistic models including DNNs. Both methods directly rely on the automatic differentiation engines
available for standard computational graphs.  In this way, the methods can be seen as extending the automatic
differentiation methods of standard computational graphs to SCGs, giving rise to a powerful approach to VI for
generic probabilistic models. The main idea underlying both approaches is to compute the gradient of the expectation given in Equation~\eqref{eq:gradELBODNNs} using Monte Carlo techniques. More precisely, we will show how we can build unbiased estimates of this gradient by sampling from the variational (or an auxiliary) distribution without having to compute the gradient of the ELBO analytically \citep{ranganath2014black,wingate2013automated,mnih2014neural}.

\vspace{-25pt}
\subsubsection{Pathwise Gradients}
\label{sec:pathwise-gradients}
The idea of this approach is to exploit reparameterizations of the variational distribution in terms of deterministic transformations of a noise distribution \citep{glasserman2013monte,fu2006gradient}. A distribution $q(\bmr|\bmnu)$ is reparameterizable if it can be expressed as
 \begin{equation}\label{eq:reparam}
\begin{split}
&\bmepsilon\sim q(\bmepsilon),\\
& \bmr = t(\bmepsilon; \bmnu),
\end{split}
\end{equation}
\noindent where $\bmepsilon$ does not depend on parameter $\bmnu$ and  $t(\cdot; \bmnu)$ is a deterministic function which encapsulates the 
dependence of $\bmr$ with respect to $\bmnu$. This transforms the expectation over  $\bmr$ to an expectation
over $\bmepsilon$. By exploiting this reparametrization property, we can express the gradient of $\lb$ in  
Equation~\eqref{eq:gradELBODNNs} as \citep{kingma2013auto,rezende2014stochastic,titsias2014doubly},

\begin{equation}\label{eq:gradELBOReparam}
\begin{split}
\nabla_{\bm \nu} {\cal L}(\bmnu) &= \nabla_{\bm \nu} \E_{\bmR} \left[ g(\bmr,\bmnu)\right]\\
&=\nabla_{\bm \nu} \E_{\bm \epsilon}\left[ g(t(\bmepsilon; \bmnu),\bmnu)\right]\\
&= \E_{\bm \epsilon}\left[  \nabla_{\bm \nu}  g(t(\bmepsilon; \bmnu),\bmnu)\right]\\
&= \E_{\bm \epsilon}\left[  \nabla_t g( t(\bmepsilon; \bmnu),\bmnu)^T\nabla_{\bm \nu} t(\bmepsilon; \bmnu) + \nabla_{\bm \nu} g( t(\bmepsilon; \bmnu),\bmnu)\right]\\
&= \E_{\bm \epsilon}\left[  \nabla_{\bm r} g(\bmr,\bmnu)^T\nabla_{\bm \nu} t(\epsilon; \bmnu) + \nabla_{\bm \nu} g(\bmr,\bmnu)\right]\\
&= \E_{\bm \epsilon}\left[  \nabla_{\bm r} g(\bmr,\bmnu)^T\nabla_{\bm \nu} t(\epsilon; \bmnu)\right] .\\
\end{split}
\end{equation}
In the last step we have exploited that $\E_{\bm \epsilon}[\nabla_{\bm \nu} g(\bmr,\bmnu)]=0$. 
To see this, we first utilize that
\[
\E_{\bm \epsilon}[\nabla_{\bm \nu} g(\bmr,\bmnu)] =  \int q(\bmepsilon) \nabla_{\bm \nu} g(\bmr,\bmnu)  d\bmepsilon =  \int q(\bmr|\bmnu) \nabla_{\bm \nu} g(\bmr,\bmnu)  d\bmr 
= \E_{\bmR}[\nabla_{\bm \nu} g(\bmr,\bmnu)].
\]
\noindent
Next, as $g(\bmr,\bmnu) = \ln p(\bmx,\bmr) - \ln q(\bmr\given\bmnu)$,  it follows that $\nabla_{\bm \nu} g(\bmr,\bmnu) = -
\nabla_{\bm \nu} \ln q(\bmr|\bmnu)$. 
Finally, since $\E_{\bmR} [\nabla_{\bm \nu} \ln q(\bmr|\bmnu)]=0$, we have that $\E_{\bm \epsilon}[\nabla_{\bm \nu} g(\bmr,\bmnu)]=0$.

Note that once we employ this reparameterization trick, the gradient enters the expectation, and afterwards we simply apply the chain rule of derivatives. 
Here it is also worth noticing that the gradient estimator is informed by the gradient with respect to $\bmg$,
which gives the direction of the maximum posterior mode (we shall return to this issue in Section~\ref{sec:score-funct-grad}).

\begin{exmp}\label{example:normal}
The Normal distribution is the best known example where this technique can be applied: A variable $ W\sim
{\cal N} (\mu, \sigma^2)$ can be reparameterized as $\epsilon\sim {\cal N} (0,1)$ and $W = \mu +
\sigma\epsilon$. So, by exploiting this re-parametrization we can compute the gradient of stochastic functions as the one defined in Figure~\ref{fig:EvaluatingStochasticCG}, i.e. compute $\nabla_\mu \E_Z[(Z-5)^2]$, where $Z\sim {\cal N} (\mu, 1)$, 
\[
\nabla_\mu \E_Z\left[(Z-5)^2\right] = \E_\epsilon\left[\nabla_\mu (\mu + \epsilon - 5)^2\right] = \E_\epsilon\left[2(\mu + \epsilon - 5)\right] = 2(\mu-5).
\]

\noindent In practice, this expectation is approximated using Monte Carlo sampling, 

\[
\nabla_\mu \E_Z[(Z-5)^2] \approx \frac{1}{K}\sum_{i=1}^K 2(\mu + \epsilon_i - 5) \quad \epsilon_i\sim {\cal N}(0,1).
\] 

\end{exmp}

\begin{figure}[htb!]

\centering

\begin{tikzpicture}

\tikzstyle{stoc}=[circle, minimum size = 10mm, thick, draw =black!80, node distance = 16mm]
\tikzstyle{deter}=[rectangle, minimum size = 10mm, thick, draw =black!80, node distance = 16mm]
\tikzstyle{input}=[minimum size = 5mm, node distance = 3mm]

\tikzstyle{connect}=[-latex, thick]

\node[stoc] (epsilon) at (0,0)  {$\bmepsilon$};
\node[deter] (t) at (2,0)  {$t$};
\node[deter] (g) at (4,0)  {$g$};

\node[input] (v) at (2,2)  {$\bmnu$};
\node[input] (x) at (4,2)  {$\bmx$};

\path (epsilon) [connect]  edge (t);
\path (t) [connect]  edge (g);

\path (v) [connect]  edge (t);
\path (v) [connect]  edge (g);

\path (x) [connect]  edge (g);

\end{tikzpicture}
\caption{\label{fig:scgreparam} Reparameterized SCG representing the ELBO function ${\cal L}(\bmnu)$.}
\end{figure}
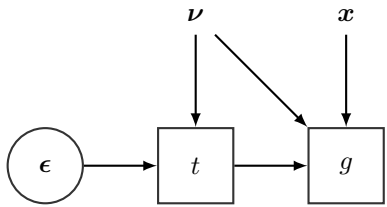

In terms of SCGs, this reparameterization trick can be captured by the transformation of the (original) SCG
shown in Figure \ref{fig:ELBOSCG} to the SCG shown in Figure \ref{fig:scgreparam}. For the transformed SCG,
the underlying CG (exemplified in Figure \ref{fig:EvaluatingStochasticCG}) can be readily applied and from
automatic differentiation we obtain unbiased estimates of the gradients of the ELBO. 

More generally, and pertinently, through the reparameterization trick we can define a CG representation of the
ELBO function ${\cal L}$, which in turn can be used for computing a Monte Carlo estimation of ${\cal L}$,

\begin{equation}
\label{eq:elboMC}
\hat{\cal L} = \frac{1}{K}\sum_{i=1}^K \ln p(\bmx,t(\bmepsilon_i;\bmnu)) - \ln q(t(\bmepsilon_i;\bmnu)|\bmnu) \quad \bmepsilon_i\sim q(\bmepsilon),
\end{equation}
\noindent and the associated automatic differentiation engine of the CG can be used for finding the derivatives of
$\calL$ (cf.\ Equation~\eqref{eq:gradELBOReparam}). The CG thus also serves as a generic tool for abstracting
away and hiding the details
of the gradient calculations from the user. 

The applicability of the reparameterization trick only extends to distributions that can be expressed in the
form shown in Equation~\eqref{eq:reparam}. 
Fortunately, \cite{figurnov2018implicit} recently introduced an \textit{implicit reparameterization approach},
which apply to a wider range of distributions including Gamma, Beta, Dirichlet and von Misses (i.e.,
distributions not covered by Equation~\eqref{eq:reparam}). This method computes the gradient of $\lb$ as  
\begin{equation}\label{eq:gradELBOReparam2}
\nabla_{\bm \nu} {\cal L}(\bmnu) = - \E_{\bmR} \left[ \frac{\nabla_{\bm r} g(\bmr,\bmnu)^T\nabla_{\bm \nu} F(\bmr; \bmnu)}{q(\bmr\given\bmnu)}\right],
\end{equation}
\noindent where $F(\bmr; \bmnu)$ is the cumulative distribution function of $q(\bmr\given\bmnu)$. Other similar approaches have been proposed for models with discrete latent random variables \citep{tucker2017rebar,grathwohl2017backpropagation}.

%

This above family of gradient estimators usually have lower variance than other methods
\citep{kucukelbir2017automatic} and,  in many cases, they can even provide good estimates with a single Monte
Carlo sample. However, the estimators only apply to distributions that support explicit or implicit
reparameterizations. Although many distributions provide this support, there are also other
relevant distributions, such as the multinomial distribution, which cannot be handled using either of the
reparameterization techniques. 




\begin{exmp}\label{example:vaeelbo}
We end this sub-section with our running example about VAEs. In this case, we consider a VAE without an encoder
network; the encoder network will be discussed in the Section~\ref{sec:elbo-optim-with}.  This model can thus
be seen as a non-linear PCA model (the non-linearity is defined in terms of an ANN) as described in Example \ref{example:vae}. For this model, the ELBO function can be expressed as

\begin{eqnarray*}
{\cal L}(\bmlambda,\bmphi) &=&\E_q[\ln p(\bmx|\bmz,\bmbeta)] +  \E_q[\ln p(\bmz)] + \E_q[\ln p(\bmbeta)]\\
&& - \E_q[\ln q(\bmz|\bmphi) -  \E_q[\ln q(\bmbeta|\bmlambda)] .
\end{eqnarray*}

Algorithm \ref{alg:vaeelbo} gives a pseudo-code specification of the SCG defining the ELBO function using
the reparameterization trick; here we only use a single sample from the variational distribution $q(\bmbeta,\bmZ|\bmlambda,\bmphi)$ in reparameterized
form. The definition of the ELBO function ${\cal L}$ is introduced together with the specification of the
decoder network, hence gradients wrt.\ the variational parameters can be readily computed and optimized using
standard algorithms.

\begin{algorithm}[htb!]
\caption{Pseudo-code for defining the ELBO function $\hat{\cal L}$, and by translation the SCG, of a VAE with no encoder network (see Algorithm \ref{alg:nonlinearPCA}). We use a single sample to compute the Monte Carlo estimate of  $\hat{\cal L}$ (see Equation~\eqref{eq:elboMC}). $\ln p_{\cal N}(\cdot|\cdot,\cdot)$ denotes the log-probability function of a Normal distribution.}\label{alg:vaeelbo}
\begin{algorithmic}
\INPUT Data: $\bmx_{train}$, Variational Parameters: $\bmlambda,\bmphi$
\STATE \# Sample (using reparameterization) from  $q(\bmbeta|\bmlambda)$ and $q(\bmz|\bmphi)$.
\STATE $\epsilon_{\alpha_0},\epsilon_{\beta_0},\epsilon_{\alpha_1}, \epsilon_{\beta_1}, \epsilon_{z} \sim {\cal N} (\bm0, \bmI)$
\STATE $\bmalpha_0 = \bmlambda_{\alpha_0,\mu} + \epsilon_{\alpha_0}\bmlambda_{\alpha_0,\sigma}, \quad \bmbeta_0 = \bmlambda_{\beta_0,\mu} + \epsilon_{\beta_0}\bmlambda_{\beta_0,\sigma}$
\STATE $\bmalpha_1 = \bmlambda_{\alpha_1,\mu} + \epsilon_{\alpha_1}\bmlambda_{\alpha_1,\sigma},\quad \bmbeta_1 = \bmlambda_{\beta_1,\mu} + \epsilon_{\beta_1}\bmlambda_{\beta_1,\sigma}$
\STATE $\bmz = \bmphi_{z,\mu} +\epsilon_z\bmphi_{z,\sigma}$ 
\STATE \# Pass the variational sample $\bmz$ through the decoder ANN 
\STATE $\bmh_0 = relu(\bmz \bmbeta_0^T + \bmalpha_0)$ 
\STATE $\bmmu_x = \bmh_0\bmbeta_1^T + \bmalpha_1$ 
\STATE \# Define the ``energy part'' of the ELBO function $E_q[\ln p(x_{train},\bmz,\bmalpha,\bmbeta)]$.
\STATE ${\cal L} = \ln p_{\cal N}(\bmx_{train}|\bmmu_x,\sigma^2_x\bmI)$
\STATE ${\cal L} = {\cal L} + \ln p_{\cal N}(\bmz|\bm0,\bmI) + \sum_i \ln p_{\cal N}(\bmalpha_i|\bm0,\bmI) + \ln p_{\cal N}(\bmbeta_i|\bm0,\bmI)$
\STATE \# Define the ``entropy part'' of the ELBO function $\E_q[\ln q(\bmz,\bmalpha,\bmbeta)]$.
\STATE ${\cal L} = {\cal L} -  \ln p_{\cal N}(\bmz|\bmphi_{z,\mu},\bmphi_{z,\sigma}^2)$
\STATE ${\cal L} = {\cal L} - \sum_i \ln p_{\cal N}(\bmalpha_i|\bmlambda_{\alpha_i,\mu},\bmlambda_{\alpha_i,\sigma}^2) 
 - \sum_i \ln p_{\cal N}(\bmbeta_i|\bmlambda_{\beta_i,\mu},\bmlambda_{\beta_i,\sigma}^2)$
\STATE return ${\cal L}$
\end{algorithmic}
\end{algorithm}
\end{exmp}

\subsubsection{Score Function Gradients}
\label{sec:score-funct-grad}
This is a classical method for gradient estimation, also known as the REINFORCE method~\citep{ranganath2014black,glynn1990likelihood,williams1992simple}. It builds on the following generic transformations to compute the gradient of an expected value, 
\begin{equation}\label{eq:gradELBOGeneral}
\begin{split}
\nabla_{\bm \nu} {\cal L}(\bmnu) &=  \nabla_{\bm \nu}\int q(\bmr\given\bmnu) g(\bmr,\bmnu) d\bmr\\
& = \int g(\bmr,\bmnu)\nabla_{\bm \nu} q(\bmr\given\bmnu)  +  q(\bmr\given\bmnu) \nabla_{\bm \nu} g(\bmr,\bmnu)d\bmr\\
& = \int g(\bmr,\bmnu)\,q(\bmr\given\bmnu) \nabla_{\bm \nu} \ln q(\bmr\given\bmnu) +  q(\bmr\given\bmnu) \nabla_{\bm \nu} g(\bmr,\bmnu)d\bmr\\
&= \E_{\bmR} \left[g(\bmr,\bmnu)\,\nabla_{\bm \nu} \ln q(\bmr\given\bmnu)  + \nabla_{\bm \nu} g(\bmr,\bmnu)\right].
\end{split}
\end{equation}
Following the discussion surrounding the derivation of Equation~\eqref{eq:gradELBOReparam}, we have that $\E_{\bmR} [\nabla_{\bm \nu}
g(\bmr,\bmnu)] =  \E_{\bmR} [-\nabla_{\bm \nu} \ln q(\bmr|\bmnu)] = 0$ and the gradient of the ELBO therefore simplifies to
\begin{equation}\label{eq:gradELBOScore}
\nabla_{\bm \nu} {\cal L}(\bmnu) = \E_{\bmR} \left[g(\bmr,\bmnu)\,\nabla_{\bm \nu}\ln q(\bmr\given\bmnu) \right].
\end{equation}
The term $\nabla_{\bm \nu} \ln q(\bmr\given\bmnu)$ (the gradient of the log of a probability distribution) is
referred to as the \emph{score function}, hence the name of the method.

From the above equation, we obtain unbiased estimates of the gradient by sampling from
$q(\bmr\given\bmnu)$. This method is general in the sense that it only requires being able to evaluate the function
$g(\bmr,\bmnu)$ and computing the score function, $\nabla_{\bm \nu}\ln
q(\bmr\given\bmnu)$. In consequence, the method applies to a wide range of models, including those covered by
the 
\textit{pathwise gradient} estimator. However, in practice, the score function gradient often yields high
variance estimates when the dimensionality of $\bmnu$ is relatively high. This is accentuated by the gradient estimator only
being guided by the gradient of the (log of the) variational distribution and not the likelihood term of the model
(which was the case for the pathwise gradient estimator). To reduce the variance, one often relies on variance
reduction techniques for improved performance
\citep{ruiz2016generalized,ranganath2014black,titsias2014doubly,mnih2016variational}, but, still, in a practical
setting the score function estimator mostly serve as the fall-back method when the pathwise gradient estimator is not applicable.

\begin{exmp}\label{example:normal2}
We revisit Example \ref{example:normal}. We have to compute the gradient of an expectation $\nabla_\mu
\E_Z[(Z-5)^2]$, where $Z\sim {\cal N} (\mu, 1)$. By applying the score function gradient estimator, we get

\begin{eqnarray*}
\nabla_\mu \E_Z[(Z-5)^2] &=& \E_Z[ (Z - 5)^2 \nabla_\mu \ln N(Z|\mu,1) ] \\
&=& \E_Z\left[ (Z - 5)^2 \nabla_\mu \left(-\frac{1}{2} (Z-\mu)^2\right)\right] \\
&=&\E_Z[ (Z - 5)^2 (Z-\mu)]  ,
\end{eqnarray*}

\noindent which can be approximated by Monte Carlo sampling, 
$
\nabla_\mu \E_Z[(Z-5)^2] \approx \frac{1}{K}\sum_{i=1}^K (z_i - 5)^2 (z_i-\mu) $, 
 where  $z_i$  are samples from ${\cal N}(\mu,1)$.

\end{exmp}


In \cite{foerster2018dice}, it is detailed an elegant implementation of this technique using SCGs.

\subsection{ELBO optimization with Amortized Variational Inference}
\label{sec:elbo-optim-with}

In principle, we can address the optimization of the ELBO function using an off-the-shelf gradient ascent
algorithm combined with the techniques presented in the previous section. The ELBO function ${\cal
  L} (\bmlambda, \bmphi)$, in this case, is again expressed  in terms of global variational parameters $\bmlambda$ (defining
the variational distribution over the global latent variables $q(\bmbeta|\bmlambda)$) and in terms of local
variational parameters $\bmphi$ (defining the variational distribution over the local latent variables
$q(\bmz_i|\bmphi_i)$); we implicitly assume that the variational posterior fully factorizes, as shown in
Equation~\eqref{eq:meanfieldQ}, although this assumption is not crucial for the discussion below.
Unfortunately, as the number of local variational parameters $\bmphi =
(\bmphi_1,\ldots,\bmphi_N)$ grows with the size  $N$ of the data set, straight-forward optimization using gradient
ascent quickly becomes computationally infeasible as the size of the data grows.

To address this issue we can rely on some of the tricks detailed in Section \ref{sec:ScalableVI}. First, we can express ${\cal L} (\bmlambda, \bmphi)$ only in terms of $\bmlambda$, as previously shown in Equations~\eqref{eq:scalable:elbo} and \eqref{eq:scalable:elboreparam},
\begin{equation*}\label{eq:amortized:elboreparam}
{\cal L}(\bmlambda) = \E_q[\ln p(\bmbeta)] - \E_q[\ln q(\bmbeta\given\bmlambda)] + \sum_{i=1}^N \max_{\bm \phi_i} (\E_q[\ln p(\bmx_i,\bmZ_i\given\bmbeta)] - \E_q[\ln q(\bmZ_i\given\bmphi_i)]) .
\end{equation*}

As done in Section \ref{sec:ScalableVI}, we can get unbiased noisy estimates of this ELBO by data
subsampling. If $I$ is a randomly chosen data index, $I\in \{1,\ldots,N\}$, and
\begin{equation*}
{\cal L}_I(\bmlambda) = \E_q[\ln p(\bmbeta)] - \E_q[\ln q(\bmbeta\given\bmlambda)] + N \max_{\bm \phi_I} (\E_q[\ln p(\bmx_I,\bmZ_I\given\bmbeta)] - \E_q[\ln q(\bmZ_I\given\bmphi_I)]) ,
\end{equation*}
then the expectation of ${\cal L}_I(\bmlambda)$ is equal to ${\cal L}(\bmlambda)$
\citep{HoffmanBleiWangPaisley13} and computing the gradient of ${\cal L}_I(\bmlambda)$ wrt.\  $\bmlambda$
will give us a noisy unbiased estimate. However, in this case, we require solving an
maximization problem for each subsampled data point (i.e. $\max_{\bm \phi_I}$). In the case of conjugate
exponential models, this inner maximization step can be computed in closed form as shown in
Equation~\eqref{eq:CoordinateAscent}. However, for models outside the conjugate exponential family, we would
have to resort to iterative algorithms, based on the  methods described in Section \ref{sec:blackbox}, making
the approach infeasible.

Amortized inference \citep{dayan1995helmholtz,gershman2014amortized} aims to address this problem by learning a mapping function, denoted by $\bms$, between $\bmx_i$ and $\bmphi_i$ parameterized by $\bmtheta$, i.e. $\bmphi_i = \bms(\bmx_i|\bmtheta)$. Hence,  ${\cal L}_I (\bmlambda)$ is expressed as   ${\cal L}_I (\bmlambda, \bmtheta)$, 
\begin{eqnarray*}
{\cal L}_I(\bmlambda, \bmtheta) &=& \E_q[\ln p(\bmbeta)] - \E_q[\ln q(\bmbeta\given\bmlambda)] \\
&+& N\cdot\E_q[\ln p(\bmx_I,\bmZ_I\given\bmbeta)] - N\cdot \E_q[\ln q(\bmZ_I\given\bmx_I,\bmphi_I = \bms(\bmx_i\given\bmtheta))] .
\end{eqnarray*}

The parameter vector $\bmtheta$ is shared among all the data points and does not grow with the data set as was
previously the case when each data point was assigned its own local variational parameters,
$\bmphi=\{\bmphi_1,\ldots,\bmphi_N\}$. On the other hand, amortized inference assumes that the parameterized
function $\bms$ is flexible enough to allow for the estimation of the local variational parameters 
$\bmphi_i$ from the data points $\bmx_i$. Thus, the family of variational distributions defined by this technique,

$$q(\bmbeta,\bmz\given\bmx,\bmlambda,\bmtheta) = q(\bmbeta\given\bmlambda)\prod_{i=1}^N q(\bmz_i\given\bmx_i, \bmphi_i=\bms(\bmx_i\given\bmtheta)) ,$$
\noindent is more restricted than the one defined in Equation~\eqref{eq:meanfieldQ}, which directly depends of $\bmlambda$ and $\bmphi$. So, there is a trade-off between flexibility in the variational approximation and computational efficiency when applying amortized inference techniques.

%

Note that the amortized function greatly simplifies the use of the model when making predictions over unseen
data $\bmx'$. If we need the posterior $p(\bmz'|\bmx')$ over a new data sample $\bmx'$ (e.g.\ for dimensionality reduction when using a VAE model), we just need to invoke the learnt amortized inference function  to recover this posterior, $q(\bmz'|\bmphi = \bms(\bmx'|\bmtheta^\star))$. 

An unbiased estimate of the gradient of ${\cal L}_I (\bmlambda, \bmtheta)$ wrt to $\bmlambda$ and $\bmtheta$ can be computed using the techniques described in the previous section, as both affect an expectation term. Note that the unbiased estimate of the gradient of ${\cal L}_I (\bmlambda, \bmtheta)$ is also an unbiased estimate of the gradient of ${\cal L} (\bmlambda, \bmtheta)$. Similarly to Equation~\eqref{eq:gradELBONoisy}, the ELBO can be maxizimed by following noisy estimates of the gradient, 
\begin{equation}\label{eq:amoritized:gradELBONoisy}
\begin{split}
\bmlambda_{t+1} = \bmlambda_t + \rho_t \hat{\nabla}_{\bm \lambda} {\cal L}_{I_t}(\bmlambda_t, \bmtheta_t),\\
\bmtheta_{t+1} = \bmtheta_t + \rho_t \hat{\nabla}_{\bm \theta} {\cal L}_{I_t}(\bmlambda_t,\bmtheta_t)
\end{split}
\end{equation}
\noindent where $I_t$ are the indexes of randomly subsampled data points at time step $t$. 


%
\begin{figure}[htb!]
	\centering
	\resizebox{\linewidth}{!}{
			\begin{tabular}{cc}
			\includegraphics{pca_nolineal_mnist.pdf}
            &
			\includegraphics{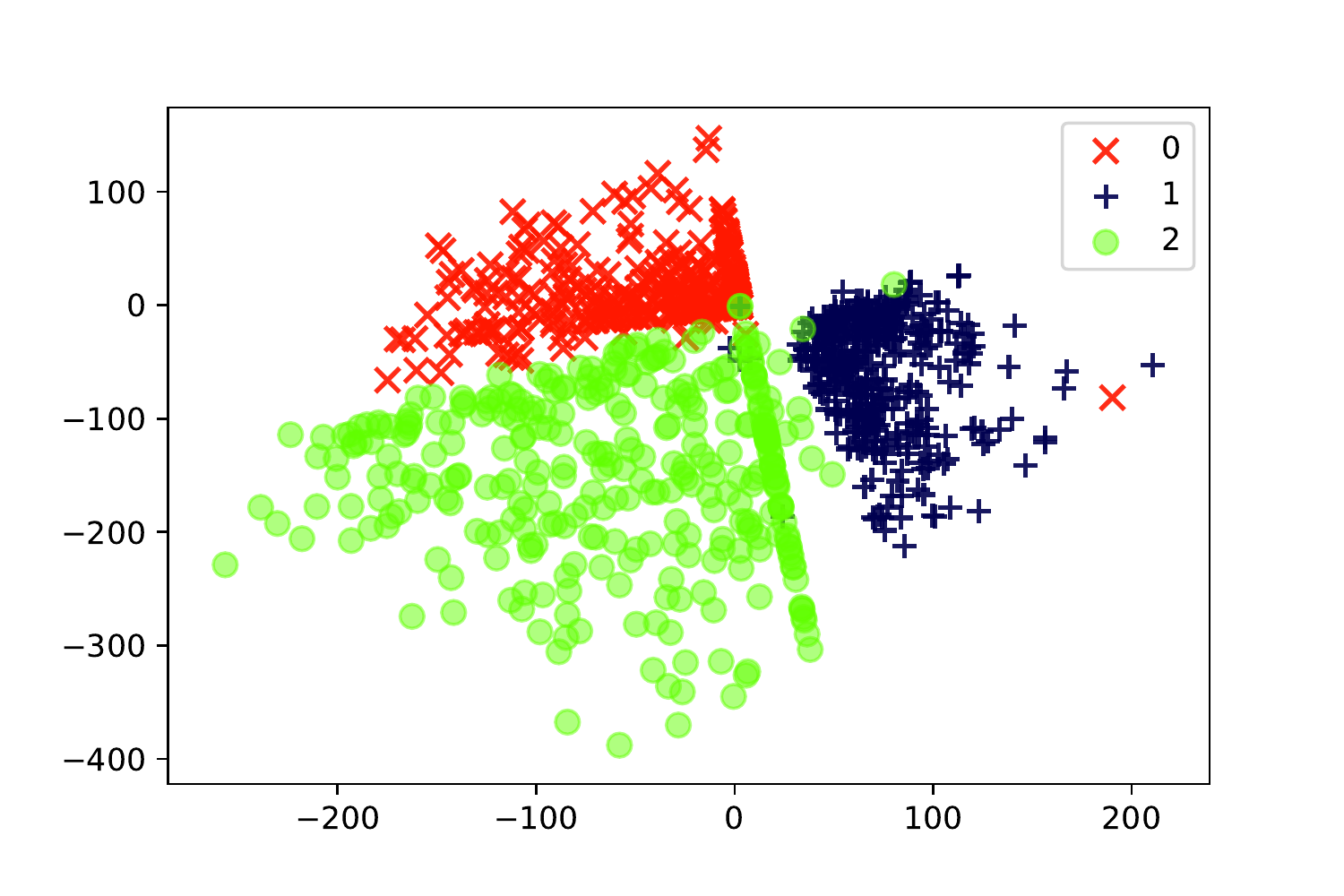}

		\end{tabular}
	}	
	\caption{\label{fig:VAE:Application} 2-dimensional latent representation of the the MNIST dataset resulting of applying: \textbf{(Left)}  a non-linear probabilistic PCA, and \textbf{(Right)} a VAE. The ANNs of the non-linear PCA and the ones defining the VAE's decoder and econder contain a single hidden layer of size 100.}
\end{figure}

\begin{exmp}\label{example:vaeelboamortized}
We finally arrive at the original formulation of VAEs, which includes an amortized inference function linking
the data samples with the latent variables $\bmZ$. This amortized function takes the form of a neural network
and is 
referred to as the \textit{encoder network} because it translates an observation $\bmX$ to (a distribution
over) its hidden representation $\bmZ$; recall that the \textit{decoder network}  (part of the model specification)
links the latent variables $\bmZ$ to (a distribution over) the observable variables $\bmX$. The existence of these two networks, the encoder and the decoder, establishes a direct link with the previously known auto-encoder networks \citep{hinton2006reducing}. In this example, both the encoder and the decoder network have a single hidden layer with a relu activation function. 

Algorithm \ref{alg:vaeelbo} shows pseudo-code defining the ELBO function associated with this model. The model
falls outside the conjugate exponential family, but due to distributional assumptions of the VAE's we can
estimate the gradients by applying the
reparameterization trick (see Section~\ref{sec:pathwise-gradients}). Specifically, from the encoder network we
sample from the variational distribution over $\bmZ_I$ given $\bmX_I$ (in reparameterized form), and at the
end of the algorithm 
we define the ELBO function ${\cal L}_I$, which includes the definition of the decoder network. As for the
previous example, the pseudo-code specification directly translates into a computational graph. From this
representation, the gradients wrt.\ the variational parameters can be readily computed and the ELBO function
optimized using, in this case, stochastic gradient ascent or some of a variation hereof.

\begin{algorithm}[htb!]
\caption{Pseudo-code for the estimation of the ELBO function ${\cal L}_I$ of a Variational Auto-encoder. We use a single sample to compute the Monte Carlo estimation of  $\hat{\cal L}$ (see Equation~\eqref{eq:elboMC}). $\ln p_{\cal N}(\cdot|\cdot,\cdot)$ denotes the log-probability function of a Normal distribuiton.}\label{alg:vaeelboamortized}
\begin{algorithmic}
\INPUT Data: $\bmx_{I}$ a single data-sample, $N$ size of the data, Variational Parameters: $\bmlambda, \bmtheta$
\STATE \# Sample (using reparametrization) from  $q(\bmbeta|\bmlambda)$.
\STATE $\epsilon_{\theta_0},\epsilon_{\theta_{0}'},\epsilon_{\theta_1}, \epsilon_{\theta_{1}'} \sim {\cal N} (\bm0, \bmI)$
\STATE $\bmtheta_{0} = \bmlambda_{\theta_0,\mu} + \epsilon_{\theta_0}\bmlambda_{\theta_0,\sigma}, \quad \bmtheta_{0}'= \bmlambda_{\theta_{0}',\mu} + \epsilon_{\theta_{0}'}\bmlambda_{\theta_{0}',\sigma}$
\STATE $\bmtheta_1 = \bmlambda_{\theta_1,\mu} + \epsilon_{\theta_1}\bmlambda_{\theta_1,\sigma},\quad \bmtheta_{1}' = \bmlambda_{\theta_{1}',\mu} + \epsilon_{\theta_{1}'}\bmlambda_{\theta_{1}',\sigma}$
\STATE \# Pass $\bmx$ through the encoder network and sample  $\bmz_I\sim q(\bmz|\bmphi = \bms(\bmx_I|\bmtheta))$
\STATE $\bmh_{z,0} = relu(\bmx_I \bmtheta_0^T + \bmtheta_0')$ 
\STATE $\bmh_{z,1} = \bmh_{z,0}\bmtheta_1^T + \bmtheta_1'$ 
\STATE \# $\bmh_{z,1}$ contains both the mean, $\bmh_{z,1,\mu}$, and the scale, $\bmh_{z,1,\sigma}$.
\STATE $\epsilon_{z} \sim {\cal N} (\bm0, \bmI)$
\STATE $\bmz_I = \bmh_{z,1,\mu} + \epsilon_{z}\bmh_{z,1,\sigma}$
\STATE \# Pass the variational sample $\bmz$ through the decoder network 
\STATE $\epsilon_{\alpha_0},\epsilon_{\beta_0},\epsilon_{\alpha_1}, \epsilon_{\beta_1} \sim {\cal N} (\bm0, \bmI)$
\STATE $\bmalpha_0 = \bmlambda_{\alpha_0,\mu} + \epsilon_{\alpha_0}\bmlambda_{\alpha_0,\sigma}, \quad \bmbeta_0 = \bmlambda_{\beta_0,\mu} + \epsilon_{\beta_0}\bmlambda_{\beta_0,\sigma}$
\STATE $\bmalpha_1 = \bmlambda_{\alpha_1,\mu} + \epsilon_{\alpha_1}\bmlambda_{\alpha_1,\sigma},\quad \bmbeta_1 = \bmlambda_{\beta_1,\mu} + \epsilon_{\beta_1}\bmlambda_{\beta_1,\sigma}$
\STATE $\bmh_0 = relu(\bmz_I \bmbeta_0^T + \bmalpha_0)$ 
\STATE $\bmmu_x = \bmh_0\bmbeta_1^T + \bmalpha_1$ 
\STATE \# Define the ``energy part'' of the ELBO function 
\STATE ${\cal L}_I = N\cdot \ln p_{\cal N}(\bmx_{I}|\bmmu_x,\sigma^2_x\bmI) + N\cdot \ln p_{\cal N}(\bmz_I|\bm0,\bmI) $
\STATE ${\cal L}_I = {\cal L}_I  + \sum_i \ln p_{\cal N}(\bmalpha_i|\bm0,\bmI) + \ln p_{\cal N}(\bmbeta_i|\bm0,\bmI)$
\STATE \# Define the ``entropy part'' of the ELBO function 
\STATE ${\cal L}_I = {\cal L}_I -  N\cdot \ln p_{\cal N}(\bmz_I|\bmh'_{1,\mu},\bmh'_{1,\sigma})$
\STATE ${\cal L}_I = {\cal L}_I - \sum_i \ln p_{\cal N}(\bmalpha_i|\bmlambda_{\alpha_i,\mu},\bmlambda_{\alpha_i,\sigma}^2) + \ln p_{\cal N}(\bmbeta_i|\bmlambda_{\beta_i,\mu},\bmlambda_{\beta_i,\sigma}^2)$
\STATE return ${\cal L}_I$
\end{algorithmic}
\end{algorithm}

%
%

\figref{VAE:Application} shows the two-dimensional latent embedding found by the non-linear probabilistic PCA (Left; reproduced from \figref{PCANonLineal:Application}) and VAE (Right) for the same reduced MNIST data set used previously. The three classes are clearly separated in latent space for both models.

\end{exmp}




%% file: conclusions.tex
\section{Conclusions and Open Issues}\label{sec:conclusions}

%
%
%

In this paper we have discussed the recent breakthroughs in approximate inference for PGMs. 
In particular, we have considered variational inference (VI), a scalable and versatile approach for doing
approximate inference in probabilistic models. The versatility of VI enables the data analyst to build
flexible models, without the constraints of limiting modelling assumptions (e.g. linear relationship between random variables).
VI is supported by a sound and well-understood mathematical foundation and exhibit good theoretical properties. 
For instance, VI is (theoretically) guaranteed to converge to an approximate posterior $q$, contained in a
set of viable approximations $\calQ$, that corresponds to a (local) maximum of the 
ELBO function, as defined in Equation \eqref{eq:elbo}. 
Nevertheless, variational inference often encounters difficulties when used in practice. 
Different random initializations of the parameter space can have significant effect on the end-result and, unless extra care is taken, issues wrt.\ numerical stability may also endanger the robustness of the obtained results. 
More research is needed to develop practical guidelines for using variational inference.

As the power of deep neural networks have entered PGMs, the PGM community has largely responded enthusiastically, embracing the new extensions to the PGM toolbox and used them eagerly. 
This has lead to new and interesting tools and models, some of which are discussed in this paper. However, we also see a potential pitfall here: 
The trend is to move away from the {modelling paradigm} that the PGM community has traditionally held in so
high regard and instead move towards catch-all LVMs (like the one depicted in \figref{plateModel}). 
These models ``\textit{let the data speak for itself}'', but at the cost of interpretability. 
PGMs are typically seen as fully transparent  models, but risk becoming more opaque with the increased
emphasis on LVMs parameterized through deep neural networks and driven by general purpose inference techniques. 
%
Initial steps have, however, already been made to leverage the PGM's modelling power 
also in this context
(e.g., \cite{johnson2016composing} combines structured latent variable representations with non-linear
likelihood functions), but a
seamless and transparent integration of neural networks and PGMs still requires further developments: 
Firstly, in a PGM model where some variables are defined using traditional probability distributions and others use deep neural networks, parts of the model may lend itself to efficient approximative inference (e.g., using VMP as described in \secref{VMP}), while others do not. 
An inference engine that utilizes an efficient (mixed) strategy approach for approximate inference in such models  would be a valuable contribution.
Secondly, VI reduces the inference problem to a continuous optimization problem. 
However, this is insufficient if the model contains latent categorical variables. 
While some PPLs, like the current release\footnote{Pyro version 0.4.1.} of Pyro, \citep{bingham2018pyro}, implements automatic enumeration over discrete latent variables, 
alternative approaches like the Concrete distribution \citep{maddison2016concrete} are also gaining some popularity. 
Thirdly, with a combined focus on inference and modeling, we may balance the results of performing approximate
inference in \lq exact models\rq and performing exact inference in \lq approximate models\rq (with the understanding that all
models are approximations).
Here, the modelling approach may lead to better understood approximations, and therefore give results that are more robust and better suited for decision support.


%% file: texmain.bbl
\begin{thebibliography}{93}
\expandafter\ifx\csname natexlab\endcsname\relax\def\natexlab#1{#1}\fi
\providecommand{\url}[1]{\texttt{#1}}
\providecommand{\href}[2]{#2}
\providecommand{\path}[1]{#1}
\providecommand{\DOIprefix}{doi:}
\providecommand{\ArXivprefix}{arXiv:}
\providecommand{\URLprefix}{URL: }
\providecommand{\Pubmedprefix}{pmid:}
\providecommand{\doi}[1]{\href{http://dx.doi.org/#1}{\path{#1}}}
\providecommand{\Pubmed}[1]{\href{pmid:#1}{\path{#1}}}
\providecommand{\bibinfo}[2]{#2}
\ifx\xfnm\undefined \def\xfnm[#1]{\unskip,\space#1}\fi
\bibitem[{Abadi et~al.(2015)Abadi, Agarwal, Barham, Brevdo, Chen, Citro,
  Corrado, Davis, Dean, Devin, Ghemawat, Goodfellow, Harp, Irving, Isard, Jia,
  Jozefowicz, Kaiser, Kudlur, Levenberg, Man\'{e}, Monga, Moore, Murray, Olah,
  Schuster, Shlens, Steiner, Sutskever, Talwar, Tucker, Vanhoucke, Vasudevan,
  Vi\'{e}gas, Vinyals, Warden, Wattenberg, Wicke, Yu and
  Zheng}]{tensorflow2015-whitepaper}
\bibinfo{author}{Abadi\xfnm[ M.]}, \bibinfo{author}{Agarwal\xfnm[ A.]},
  \bibinfo{author}{Barham\xfnm[ P.]}, \bibinfo{author}{Brevdo\xfnm[ E.]},
  \bibinfo{author}{Chen\xfnm[ Z.]}, \bibinfo{author}{Citro\xfnm[ C.]},
  \bibinfo{author}{Corrado\xfnm[ G.S.]}, \bibinfo{author}{Davis\xfnm[ A.]},
  \bibinfo{author}{Dean\xfnm[ J.]}, \bibinfo{author}{Devin\xfnm[ M.]},
  \bibinfo{author}{Ghemawat\xfnm[ S.]}, \bibinfo{author}{Goodfellow\xfnm[ I.]},
  \bibinfo{author}{Harp\xfnm[ A.]}, \bibinfo{author}{Irving\xfnm[ G.]},
  \bibinfo{author}{Isard\xfnm[ M.]}, \bibinfo{author}{Jia\xfnm[ Y.]},
  \bibinfo{author}{Jozefowicz\xfnm[ R.]}, \bibinfo{author}{Kaiser\xfnm[ L.]},
  \bibinfo{author}{Kudlur\xfnm[ M.]}, \bibinfo{author}{Levenberg\xfnm[ J.]},
  \bibinfo{author}{Man\'{e}\xfnm[ D.]}, \bibinfo{author}{Monga\xfnm[ R.]},
  \bibinfo{author}{Moore\xfnm[ S.]}, \bibinfo{author}{Murray\xfnm[ D.]},
  \bibinfo{author}{Olah\xfnm[ C.]}, \bibinfo{author}{Schuster\xfnm[ M.]},
  \bibinfo{author}{Shlens\xfnm[ J.]}, \bibinfo{author}{Steiner\xfnm[ B.]},
  \bibinfo{author}{Sutskever\xfnm[ I.]}, \bibinfo{author}{Talwar\xfnm[ K.]},
  \bibinfo{author}{Tucker\xfnm[ P.]}, \bibinfo{author}{Vanhoucke\xfnm[ V.]},
  \bibinfo{author}{Vasudevan\xfnm[ V.]}, \bibinfo{author}{Vi\'{e}gas\xfnm[
  F.]}, \bibinfo{author}{Vinyals\xfnm[ O.]}, \bibinfo{author}{Warden\xfnm[
  P.]}, \bibinfo{author}{Wattenberg\xfnm[ M.]}, \bibinfo{author}{Wicke\xfnm[
  M.]}, \bibinfo{author}{Yu\xfnm[ Y.]}, \bibinfo{author}{Zheng\xfnm[ X.]}.
\newblock \bibinfo{title}{{TensorFlow}: Large-scale machine learning on
  heterogeneous systems}.
\newblock \bibinfo{year}{2015}.
\newblock \URLprefix \url{http://tensorflow.org/}; \bibinfo{note}{software
  available from tensorflow.org}.
\bibitem[{Amari(1998)}]{amari1998natural}
\bibinfo{author}{Amari\xfnm[ S.I.]}.
\newblock \bibinfo{title}{Natural gradient works efficiently in learning}.
\newblock \bibinfo{journal}{Neural Computation}
  \bibinfo{year}{1998};\bibinfo{volume}{10}(\bibinfo{number}{2}):\bibinfo{pages}{251--276}.
\bibitem[{Barndorff-Nielsen(2014)}]{barndorff2014information}
\bibinfo{author}{Barndorff-Nielsen\xfnm[ O.]}.
\newblock \bibinfo{title}{Information and exponential families in statistical
  theory}.
\newblock \bibinfo{publisher}{John Wiley \& Sons}, \bibinfo{year}{2014}.
\bibitem[{Bergstra et~al.(2010)Bergstra, Breuleux, Bastien, Lamblin, Pascanu,
  Desjardins, Turian, Warde-Farley and Bengio}]{bergstra2010theano}
\bibinfo{author}{Bergstra\xfnm[ J.]}, \bibinfo{author}{Breuleux\xfnm[ O.]},
  \bibinfo{author}{Bastien\xfnm[ F.]}, \bibinfo{author}{Lamblin\xfnm[ P.]},
  \bibinfo{author}{Pascanu\xfnm[ R.]}, \bibinfo{author}{Desjardins\xfnm[ G.]},
  \bibinfo{author}{Turian\xfnm[ J.]}, \bibinfo{author}{Warde-Farley\xfnm[ D.]},
  \bibinfo{author}{Bengio\xfnm[ Y.]}.
\newblock \bibinfo{title}{Theano: a {CPU} and {GPU} math expression compiler}.
\newblock In: \bibinfo{booktitle}{Proceedings of the Python for scientific
  computing conference (SciPy)}. \bibinfo{organization}{Austin, TX};
  volume~\bibinfo{volume}{4}; \bibinfo{year}{2010}. p. \bibinfo{pages}{3--10}.
\bibitem[{Bingham et~al.(2018)Bingham, Chen, Jankowiak, Obermeyer, Pradhan,
  Karaletsos, Singh, Szerlip, Horsfall and Goodman}]{bingham2018pyro}
\bibinfo{author}{Bingham\xfnm[ E.]}, \bibinfo{author}{Chen\xfnm[ J.P.]},
  \bibinfo{author}{Jankowiak\xfnm[ M.]}, \bibinfo{author}{Obermeyer\xfnm[ F.]},
  \bibinfo{author}{Pradhan\xfnm[ N.]}, \bibinfo{author}{Karaletsos\xfnm[ T.]},
  \bibinfo{author}{Singh\xfnm[ R.]}, \bibinfo{author}{Szerlip\xfnm[ P.]},
  \bibinfo{author}{Horsfall\xfnm[ P.]}, \bibinfo{author}{Goodman\xfnm[ N.D.]}.
\newblock \bibinfo{title}{Pyro: Deep universal probabilistic programming}.
\newblock \bibinfo{journal}{CoRR}
  \bibinfo{year}{2018};\href{http://arxiv.org/abs/1810.09538}{\tt
  arXiv:1810.09538}.
\bibitem[{Bishop(1998)}]{bishop1998latent}
\bibinfo{author}{Bishop\xfnm[ C.M.]}.
\newblock \bibinfo{title}{Latent variable models}.
\newblock In: \bibinfo{booktitle}{Learning in graphical models}.
  \bibinfo{publisher}{Springer}; \bibinfo{year}{1998}. p.
  \bibinfo{pages}{371--403}.
\bibitem[{Bishop(2006)}]{bishop2006pattern}
\bibinfo{author}{Bishop\xfnm[ C.M.]}.
\newblock \bibinfo{title}{Pattern Recognition and Machine Learning}.
\newblock \bibinfo{publisher}{Springer}, \bibinfo{year}{2006}.
\bibitem[{Blei(2014)}]{blei2014build}
\bibinfo{author}{Blei\xfnm[ D.M.]}.
\newblock \bibinfo{title}{Build, compute, critique, repeat: Data analysis with
  latent variable models}.
\newblock \bibinfo{journal}{Annual Review of Statistics and its Application}
  \bibinfo{year}{2014};\bibinfo{volume}{1}:\bibinfo{pages}{203--232}.
\bibitem[{Blei et~al.(2003)Blei, Ng and Jordan}]{blei2003latent}
\bibinfo{author}{Blei\xfnm[ D.M.]}, \bibinfo{author}{Ng\xfnm[ A.Y.]},
  \bibinfo{author}{Jordan\xfnm[ M.I.]}.
\newblock \bibinfo{title}{Latent {D}irichlet allocation}.
\newblock \bibinfo{journal}{Journal of Machine Learning Research}
  \bibinfo{year}{2003};\bibinfo{volume}{3}(\bibinfo{number}{Jan}):\bibinfo{pages}{993--1022}.
\bibitem[{Bottou(2010)}]{bottou2010large}
\bibinfo{author}{Bottou\xfnm[ L.]}.
\newblock \bibinfo{title}{Large-scale machine learning with stochastic gradient
  descent}.
\newblock In: \bibinfo{booktitle}{Proceedings of COMPSTAT'2010}.
  \bibinfo{publisher}{Springer}; \bibinfo{year}{2010}. p.
  \bibinfo{pages}{177--186}.
\bibitem[{Caba{\~n}as et~al.(2019)Caba{\~n}as, Salmer{\'o}n and
  Masegosa}]{cabanasInferPy}
\bibinfo{author}{Caba{\~n}as\xfnm[ R.]}, \bibinfo{author}{Salmer{\'o}n\xfnm[
  A.]}, \bibinfo{author}{Masegosa\xfnm[ A.R.]}.
\newblock \bibinfo{title}{{InferPy}: Probabilistic modeling with {TensorFlow}
  made easy}.
\newblock \bibinfo{journal}{Knowledge-Based Systems}
  \bibinfo{year}{2019};\bibinfo{volume}{168}:\bibinfo{pages}{25--27}.
\bibitem[{Card et~al.(2017)Card, Tan and Smith}]{card2017neural}
\bibinfo{author}{Card\xfnm[ D.]}, \bibinfo{author}{Tan\xfnm[ C.]},
  \bibinfo{author}{Smith\xfnm[ N.A.]}.
\newblock \bibinfo{title}{A neural framework for generalized topic models}.
\newblock \bibinfo{journal}{arXiv preprint arXiv:170509296}
  \bibinfo{year}{2017};.
\bibitem[{Chen et~al.(2015)Chen, Li, Li, Lin, Wang, Wang, Xiao, Xu, Zhang and
  Zhang}]{chen2015mxnet}
\bibinfo{author}{Chen\xfnm[ T.]}, \bibinfo{author}{Li\xfnm[ M.]},
  \bibinfo{author}{Li\xfnm[ Y.]}, \bibinfo{author}{Lin\xfnm[ M.]},
  \bibinfo{author}{Wang\xfnm[ N.]}, \bibinfo{author}{Wang\xfnm[ M.]},
  \bibinfo{author}{Xiao\xfnm[ T.]}, \bibinfo{author}{Xu\xfnm[ B.]},
  \bibinfo{author}{Zhang\xfnm[ C.]}, \bibinfo{author}{Zhang\xfnm[ Z.]}.
\newblock \bibinfo{title}{Mxnet: A flexible and efficient machine learning
  library for heterogeneous distributed systems}.
\newblock \bibinfo{journal}{arXiv preprint arXiv:151201274}
  \bibinfo{year}{2015};.
\bibitem[{Chung et~al.(2015)Chung, Kastner, Dinh, Goel, Courville and
  Bengio}]{chung2015recurrent}
\bibinfo{author}{Chung\xfnm[ J.]}, \bibinfo{author}{Kastner\xfnm[ K.]},
  \bibinfo{author}{Dinh\xfnm[ L.]}, \bibinfo{author}{Goel\xfnm[ K.]},
  \bibinfo{author}{Courville\xfnm[ A.C.]}, \bibinfo{author}{Bengio\xfnm[ Y.]}.
\newblock \bibinfo{title}{A recurrent latent variable model for sequential
  data}.
\newblock In: \bibinfo{booktitle}{Advances in Neural Information Processing
  Systems}. \bibinfo{year}{2015}. p. \bibinfo{pages}{2980--2988}.
\bibitem[{C{\'o}zar et~al.(2019)C{\'o}zar, Caba{\~n}as, Salmer{\'o}n and
  Masegosa}]{cozar2019inferpy}
\bibinfo{author}{C{\'o}zar\xfnm[ J.]}, \bibinfo{author}{Caba{\~n}as\xfnm[ R.]},
  \bibinfo{author}{Salmer{\'o}n\xfnm[ A.]}, \bibinfo{author}{Masegosa\xfnm[
  A.R.]}.
\newblock \bibinfo{title}{{InferPy}: Probabilistic modeling with deep neural
  networks made easy}.
\newblock \bibinfo{journal}{arXiv preprint arXiv:190811161}
  \bibinfo{year}{2019};.
\bibitem[{Dayan et~al.(1995)Dayan, Hinton, Neal and Zemel}]{dayan1995helmholtz}
\bibinfo{author}{Dayan\xfnm[ P.]}, \bibinfo{author}{Hinton\xfnm[ G.E.]},
  \bibinfo{author}{Neal\xfnm[ R.M.]}, \bibinfo{author}{Zemel\xfnm[ R.S.]}.
\newblock \bibinfo{title}{The {H}elmholtz machine}.
\newblock \bibinfo{journal}{Neural Computation}
  \bibinfo{year}{1995};\bibinfo{volume}{7}(\bibinfo{number}{5}):\bibinfo{pages}{889--904}.
\bibitem[{Dillon et~al.(2017)Dillon, Langmore, Tran, Brevdo, Vasudevan, Moore,
  Patton, Alemi, Hoffman and Saurous}]{dillon2017tensorflow}
\bibinfo{author}{Dillon\xfnm[ J.V.]}, \bibinfo{author}{Langmore\xfnm[ I.]},
  \bibinfo{author}{Tran\xfnm[ D.]}, \bibinfo{author}{Brevdo\xfnm[ E.]},
  \bibinfo{author}{Vasudevan\xfnm[ S.]}, \bibinfo{author}{Moore\xfnm[ D.]},
  \bibinfo{author}{Patton\xfnm[ B.]}, \bibinfo{author}{Alemi\xfnm[ A.]},
  \bibinfo{author}{Hoffman\xfnm[ M.]}, \bibinfo{author}{Saurous\xfnm[ R.A.]}.
\newblock \bibinfo{title}{{TensorFlow} distributions}.
\newblock \bibinfo{journal}{arXiv preprint arXiv:171110604}
  \bibinfo{year}{2017};.
\bibitem[{Doersch(2016)}]{doersch2016tutorial}
\bibinfo{author}{Doersch\xfnm[ C.]}.
\newblock \bibinfo{title}{Tutorial on variational autoencoders}.
\newblock \bibinfo{journal}{arXiv preprint arXiv:160605908}
  \bibinfo{year}{2016};.
\bibitem[{Figurnov et~al.(2018)Figurnov, Mohamed and
  Mnih}]{figurnov2018implicit}
\bibinfo{author}{Figurnov\xfnm[ M.]}, \bibinfo{author}{Mohamed\xfnm[ S.]},
  \bibinfo{author}{Mnih\xfnm[ A.]}.
\newblock \bibinfo{title}{Implicit reparameterization gradients}.
\newblock \bibinfo{journal}{arXiv preprint arXiv:180508498}
  \bibinfo{year}{2018};.
\bibitem[{Fisher(1936)}]{fisher1936use}
\bibinfo{author}{Fisher\xfnm[ R.A.]}.
\newblock \bibinfo{title}{The use of multiple measurements in taxonomic
  problems}.
\newblock \bibinfo{journal}{Annals of Eugenics}
  \bibinfo{year}{1936};\bibinfo{volume}{7}(\bibinfo{number}{2}):\bibinfo{pages}{179--188}.
\bibitem[{Foerster et~al.(2018)Foerster, Farquhar, Al-Shedivat,
  Rockt{\"a}schel, Xing and Whiteson}]{foerster2018dice}
\bibinfo{author}{Foerster\xfnm[ J.]}, \bibinfo{author}{Farquhar\xfnm[ G.]},
  \bibinfo{author}{Al-Shedivat\xfnm[ M.]},
  \bibinfo{author}{Rockt{\"a}schel\xfnm[ T.]}, \bibinfo{author}{Xing\xfnm[
  E.P.]}, \bibinfo{author}{Whiteson\xfnm[ S.]}.
\newblock \bibinfo{title}{Dice: The infinitely differentiable {M}onte-{C}arlo
  estimator}.
\newblock \bibinfo{journal}{arXiv preprint arXiv:180205098}
  \bibinfo{year}{2018};.
\bibitem[{Fu(2006)}]{fu2006gradient}
\bibinfo{author}{Fu\xfnm[ M.C.]}.
\newblock \bibinfo{title}{Gradient estimation}.
\newblock \bibinfo{journal}{Handbooks in operations research and management
  science} \bibinfo{year}{2006};\bibinfo{volume}{13}:\bibinfo{pages}{575--616}.
\bibitem[{Gershman and Goodman(2014)}]{gershman2014amortized}
\bibinfo{author}{Gershman\xfnm[ S.]}, \bibinfo{author}{Goodman\xfnm[ N.]}.
\newblock \bibinfo{title}{Amortized inference in probabilistic reasoning}.
\newblock In: \bibinfo{booktitle}{Proceedings of the Annual Meeting of the
  Cognitive Science Society}. volume~\bibinfo{volume}{36};
  \bibinfo{year}{2014}. p. \bibinfo{pages}{517--522}.
\bibitem[{Ghahramani(2015)}]{ghahramani2015probabilistic}
\bibinfo{author}{Ghahramani\xfnm[ Z.]}.
\newblock \bibinfo{title}{Probabilistic machine learning and artificial
  intelligence}.
\newblock \bibinfo{journal}{Nature}
  \bibinfo{year}{2015};\bibinfo{volume}{521}(\bibinfo{number}{7553}):\bibinfo{pages}{452}.
\bibitem[{Gilks et~al.(1995)Gilks, Richardson and
  Spiegelhalter}]{gilks1995markov}
\bibinfo{author}{Gilks\xfnm[ W.R.]}, \bibinfo{author}{Richardson\xfnm[ S.]},
  \bibinfo{author}{Spiegelhalter\xfnm[ D.]}.
\newblock \bibinfo{title}{Markov Chain {M}onte {C}arlo in practice}.
\newblock \bibinfo{publisher}{Chapman and Hall/CRC}, \bibinfo{year}{1995}.
\bibitem[{Glasserman(2013)}]{glasserman2013monte}
\bibinfo{author}{Glasserman\xfnm[ P.]}.
\newblock \bibinfo{title}{Monte {C}arlo methods in financial engineering}.
\newblock volume~\bibinfo{volume}{53}.
\newblock \bibinfo{publisher}{Springer}, \bibinfo{year}{2013}.
\bibitem[{Glynn(1990)}]{glynn1990likelihood}
\bibinfo{author}{Glynn\xfnm[ P.W.]}.
\newblock \bibinfo{title}{Likelihood ratio gradient estimation for stochastic
  systems}.
\newblock \bibinfo{journal}{Communications of the ACM}
  \bibinfo{year}{1990};\bibinfo{volume}{33}(\bibinfo{number}{10}):\bibinfo{pages}{75--84}.
\bibitem[{G{\'o}mez-Bombarelli et~al.(2018)G{\'o}mez-Bombarelli, Wei, Duvenaud,
  Hern{\'a}ndez-Lobato, S{\'a}nchez-Lengeling, Sheberla, Aguilera-Iparraguirre,
  Hirzel, Adams and Aspuru-Guzik}]{gomez2018automatic}
\bibinfo{author}{G{\'o}mez-Bombarelli\xfnm[ R.]}, \bibinfo{author}{Wei\xfnm[
  J.N.]}, \bibinfo{author}{Duvenaud\xfnm[ D.]},
  \bibinfo{author}{Hern{\'a}ndez-Lobato\xfnm[ J.M.]},
  \bibinfo{author}{S{\'a}nchez-Lengeling\xfnm[ B.]},
  \bibinfo{author}{Sheberla\xfnm[ D.]},
  \bibinfo{author}{Aguilera-Iparraguirre\xfnm[ J.]},
  \bibinfo{author}{Hirzel\xfnm[ T.D.]}, \bibinfo{author}{Adams\xfnm[ R.P.]},
  \bibinfo{author}{Aspuru-Guzik\xfnm[ A.]}.
\newblock \bibinfo{title}{Automatic chemical design using a data-driven
  continuous representation of molecules}.
\newblock \bibinfo{journal}{ACS Central Science}
  \bibinfo{year}{2018};\bibinfo{volume}{4}(\bibinfo{number}{2}):\bibinfo{pages}{268--276}.
\bibitem[{Goodfellow et~al.(2016)Goodfellow, Bengio, Courville and
  Bengio}]{goodfellow2016deep}
\bibinfo{author}{Goodfellow\xfnm[ I.]}, \bibinfo{author}{Bengio\xfnm[ Y.]},
  \bibinfo{author}{Courville\xfnm[ A.]}, \bibinfo{author}{Bengio\xfnm[ Y.]}.
\newblock \bibinfo{title}{Deep Learning}.
\newblock \bibinfo{publisher}{MIT Press}, \bibinfo{year}{2016}.
\bibitem[{Goodfellow et~al.(2014)Goodfellow, Pouget-Abadie, Mirza, Xu,
  Warde-Farley, Ozair, Courville and Bengio}]{goodfellow2014generative}
\bibinfo{author}{Goodfellow\xfnm[ I.]}, \bibinfo{author}{Pouget-Abadie\xfnm[
  J.]}, \bibinfo{author}{Mirza\xfnm[ M.]}, \bibinfo{author}{Xu\xfnm[ B.]},
  \bibinfo{author}{Warde-Farley\xfnm[ D.]}, \bibinfo{author}{Ozair\xfnm[ S.]},
  \bibinfo{author}{Courville\xfnm[ A.]}, \bibinfo{author}{Bengio\xfnm[ Y.]}.
\newblock \bibinfo{title}{Generative adversarial nets}.
\newblock In: \bibinfo{booktitle}{Advances in Neural Information Processing
  Systems}. \bibinfo{year}{2014}. p. \bibinfo{pages}{2672--2680}.
\bibitem[{Gordon et~al.(2014)Gordon, Henzinger, Nori and
  Rajamani}]{gordon2014probabilistic}
\bibinfo{author}{Gordon\xfnm[ A.D.]}, \bibinfo{author}{Henzinger\xfnm[ T.A.]},
  \bibinfo{author}{Nori\xfnm[ A.V.]}, \bibinfo{author}{Rajamani\xfnm[ S.K.]}.
\newblock \bibinfo{title}{Probabilistic programming}.
\newblock In: \bibinfo{booktitle}{Proceedings of the on Future of Software
  Engineering}. \bibinfo{organization}{ACM}; \bibinfo{year}{2014}. p.
  \bibinfo{pages}{167--181}.
\bibitem[{Grathwohl et~al.(2017)Grathwohl, Choi, Wu, Roeder and
  Duvenaud}]{grathwohl2017backpropagation}
\bibinfo{author}{Grathwohl\xfnm[ W.]}, \bibinfo{author}{Choi\xfnm[ D.]},
  \bibinfo{author}{Wu\xfnm[ Y.]}, \bibinfo{author}{Roeder\xfnm[ G.]},
  \bibinfo{author}{Duvenaud\xfnm[ D.]}.
\newblock \bibinfo{title}{Backpropagation through the void: Optimizing control
  variates for black-box gradient estimation}.
\newblock \bibinfo{journal}{arXiv preprint arXiv:171100123}
  \bibinfo{year}{2017};.
\bibitem[{Gregor et~al.(2015)Gregor, Danihelka, Graves, Rezende and
  Wierstra}]{gregor2015draw}
\bibinfo{author}{Gregor\xfnm[ K.]}, \bibinfo{author}{Danihelka\xfnm[ I.]},
  \bibinfo{author}{Graves\xfnm[ A.]}, \bibinfo{author}{Rezende\xfnm[ D.J.]},
  \bibinfo{author}{Wierstra\xfnm[ D.]}.
\newblock \bibinfo{title}{Draw: A recurrent neural network for image
  generation}.
\newblock \bibinfo{journal}{arXiv preprint arXiv:150204623}
  \bibinfo{year}{2015};.
\bibitem[{Griewank(1989)}]{griewank1989automatic}
\bibinfo{author}{Griewank\xfnm[ A.]}.
\newblock \bibinfo{title}{On automatic differentiation}.
\newblock \bibinfo{journal}{Mathematical Programming: Recent Developments and
  Applications}
  \bibinfo{year}{1989};\bibinfo{volume}{6}(\bibinfo{number}{6}):\bibinfo{pages}{83--107}.
\bibitem[{Hastie et~al.(2001)Hastie, Tibshirani and
  Friedman}]{HastieTibshiraniFriedman01}
\bibinfo{author}{Hastie\xfnm[ T.]}, \bibinfo{author}{Tibshirani\xfnm[ R.]},
  \bibinfo{author}{Friedman\xfnm[ J.]}.
\newblock \bibinfo{title}{The Elements of Statistical Learning}.
\newblock \bibinfo{publisher}{Springer}, \bibinfo{year}{2001}.
\bibitem[{Hinton(2009)}]{hinton2009deep}
\bibinfo{author}{Hinton\xfnm[ G.E.]}.
\newblock \bibinfo{title}{Deep belief networks}.
\newblock \bibinfo{journal}{Scholarpedia}
  \bibinfo{year}{2009};\bibinfo{volume}{4}(\bibinfo{number}{5}):\bibinfo{pages}{5947}.
\bibitem[{Hinton(2012)}]{hinton2012practical}
\bibinfo{author}{Hinton\xfnm[ G.E.]}.
\newblock \bibinfo{title}{A practical guide to training restricted {B}oltzmann
  machines}.
\newblock In: \bibinfo{booktitle}{Neural networks: Tricks of the trade}.
  \bibinfo{publisher}{Springer}; \bibinfo{year}{2012}. p.
  \bibinfo{pages}{599--619}.
\bibitem[{Hinton and Salakhutdinov(2006)}]{hinton2006reducing}
\bibinfo{author}{Hinton\xfnm[ G.E.]}, \bibinfo{author}{Salakhutdinov\xfnm[
  R.R.]}.
\newblock \bibinfo{title}{Reducing the dimensionality of data with neural
  networks}.
\newblock \bibinfo{journal}{Science}
  \bibinfo{year}{2006};\bibinfo{volume}{313}(\bibinfo{number}{5786}):\bibinfo{pages}{504--507}.
\bibitem[{Hoffman et~al.(2013)Hoffman, Blei, Wang and
  Paisley}]{HoffmanBleiWangPaisley13}
\bibinfo{author}{Hoffman\xfnm[ M.D.]}, \bibinfo{author}{Blei\xfnm[ D.M.]},
  \bibinfo{author}{Wang\xfnm[ C.]}, \bibinfo{author}{Paisley\xfnm[ J.]}.
\newblock \bibinfo{title}{Stochastic variational inference}.
\newblock \bibinfo{journal}{Journal of Machine Learning Research}
  \bibinfo{year}{2013};\bibinfo{volume}{14}:\bibinfo{pages}{1303--1347}.
\bibitem[{Hopfield(1988)}]{hopfield1988artificial}
\bibinfo{author}{Hopfield\xfnm[ J.J.]}.
\newblock \bibinfo{title}{Artificial neural networks}.
\newblock \bibinfo{journal}{IEEE Circuits and Devices Magazine}
  \bibinfo{year}{1988};\bibinfo{volume}{4}(\bibinfo{number}{5}):\bibinfo{pages}{3--10}.
\bibitem[{Hsu et~al.(2017)Hsu, Zhang and Glass}]{hsu2017learning}
\bibinfo{author}{Hsu\xfnm[ W.N.]}, \bibinfo{author}{Zhang\xfnm[ Y.]},
  \bibinfo{author}{Glass\xfnm[ J.]}.
\newblock \bibinfo{title}{Learning latent representations for speech generation
  and transformation}.
\newblock \bibinfo{journal}{arXiv preprint arXiv:170404222}
  \bibinfo{year}{2017};.
\bibitem[{Jensen and Nielsen(2007)}]{JensenNielsen07}
\bibinfo{author}{Jensen\xfnm[ F.V.]}, \bibinfo{author}{Nielsen\xfnm[ T.D.]}.
\newblock \bibinfo{title}{{B}ayesian Networks and Decision Graphs}.
\newblock \bibinfo{address}{Berlin, Germany}: \bibinfo{publisher}{Springer},
  \bibinfo{year}{2007}.
\bibitem[{Jiang et~al.(2016)Jiang, Zheng, Tan, Tang and
  Zhou}]{jiang2016variational}
\bibinfo{author}{Jiang\xfnm[ Z.]}, \bibinfo{author}{Zheng\xfnm[ Y.]},
  \bibinfo{author}{Tan\xfnm[ H.]}, \bibinfo{author}{Tang\xfnm[ B.]},
  \bibinfo{author}{Zhou\xfnm[ H.]}.
\newblock \bibinfo{title}{Variational deep embedding: An unsupervised and
  generative approach to clustering}.
\newblock \bibinfo{journal}{arXiv preprint arXiv:161105148}
  \bibinfo{year}{2016};.
\bibitem[{Johnson et~al.(2016)Johnson, Duvenaud, Wiltschko, Adams and
  Datta}]{johnson2016composing}
\bibinfo{author}{Johnson\xfnm[ M.]}, \bibinfo{author}{Duvenaud\xfnm[ D.K.]},
  \bibinfo{author}{Wiltschko\xfnm[ A.]}, \bibinfo{author}{Adams\xfnm[ R.P.]},
  \bibinfo{author}{Datta\xfnm[ S.R.]}.
\newblock \bibinfo{title}{Composing graphical models with neural networks for
  structured representations and fast inference}.
\newblock In: \bibinfo{booktitle}{Advances in Neural Information Processing
  Systems}. \bibinfo{year}{2016}. p. \bibinfo{pages}{2946--2954}.
\bibitem[{Jordan et~al.(1999)Jordan, Ghahramani, Jaakkola and
  Saul}]{jordan1999introduction}
\bibinfo{author}{Jordan\xfnm[ M.I.]}, \bibinfo{author}{Ghahramani\xfnm[ Z.]},
  \bibinfo{author}{Jaakkola\xfnm[ T.S.]}, \bibinfo{author}{Saul\xfnm[ L.K.]}.
\newblock \bibinfo{title}{An introduction to variational methods for graphical
  models}.
\newblock \bibinfo{journal}{Machine Learning}
  \bibinfo{year}{1999};\bibinfo{volume}{37}(\bibinfo{number}{2}):\bibinfo{pages}{183--233}.
\bibitem[{Kingma and Welling(2013)}]{kingma2013auto}
\bibinfo{author}{Kingma\xfnm[ D.P.]}, \bibinfo{author}{Welling\xfnm[ M.]}.
\newblock \bibinfo{title}{Auto-encoding variational {B}ayes}.
\newblock \bibinfo{journal}{arXiv preprint arXiv:13126114}
  \bibinfo{year}{2013};.
\bibitem[{Kipf and Welling(2016)}]{kipf2016variational}
\bibinfo{author}{Kipf\xfnm[ T.N.]}, \bibinfo{author}{Welling\xfnm[ M.]}.
\newblock \bibinfo{title}{Variational graph auto-encoders}.
\newblock \bibinfo{journal}{arXiv preprint arXiv:161107308}
  \bibinfo{year}{2016};.
\bibitem[{Koller and Friedman(2009)}]{koller2009probabilistic}
\bibinfo{author}{Koller\xfnm[ D.]}, \bibinfo{author}{Friedman\xfnm[ N.]}.
\newblock \bibinfo{title}{Probabilistic graphical models: principles and
  techniques}.
\newblock \bibinfo{publisher}{MIT Press}, \bibinfo{year}{2009}.
\bibitem[{Kucukelbir et~al.(2017)Kucukelbir, Tran, Ranganath, Gelman and
  Blei}]{kucukelbir2017automatic}
\bibinfo{author}{Kucukelbir\xfnm[ A.]}, \bibinfo{author}{Tran\xfnm[ D.]},
  \bibinfo{author}{Ranganath\xfnm[ R.]}, \bibinfo{author}{Gelman\xfnm[ A.]},
  \bibinfo{author}{Blei\xfnm[ D.M.]}.
\newblock \bibinfo{title}{Automatic differentiation variational inference}.
\newblock \bibinfo{journal}{Journal of Machine Learning Research}
  \bibinfo{year}{2017};\bibinfo{volume}{18}(\bibinfo{number}{1}):\bibinfo{pages}{430--474}.
\bibitem[{Kulkarni et~al.(2015)Kulkarni, Whitney, Kohli and
  Tenenbaum}]{kulkarni2015deep}
\bibinfo{author}{Kulkarni\xfnm[ T.D.]}, \bibinfo{author}{Whitney\xfnm[ W.F.]},
  \bibinfo{author}{Kohli\xfnm[ P.]}, \bibinfo{author}{Tenenbaum\xfnm[ J.]}.
\newblock \bibinfo{title}{Deep convolutional inverse graphics network}.
\newblock In: \bibinfo{booktitle}{Advances in Neural Information Processing
  Systems}. \bibinfo{year}{2015}. p. \bibinfo{pages}{2539--2547}.
\bibitem[{Lauritzen(1992)}]{lauritzen1992propagation}
\bibinfo{author}{Lauritzen\xfnm[ S.L.]}.
\newblock \bibinfo{title}{Propagation of probabilities, means, and variances in
  mixed graphical association models}.
\newblock \bibinfo{journal}{Journal of the American Statistical Association}
  \bibinfo{year}{1992};\bibinfo{volume}{87}(\bibinfo{number}{420}):\bibinfo{pages}{1098--1108}.
\bibitem[{LeCun et~al.(1998)LeCun, Bottou, Bengio and
  Haffner}]{lecun1998gradient}
\bibinfo{author}{LeCun\xfnm[ Y.]}, \bibinfo{author}{Bottou\xfnm[ L.]},
  \bibinfo{author}{Bengio\xfnm[ Y.]}, \bibinfo{author}{Haffner\xfnm[ P.]}.
\newblock \bibinfo{title}{Gradient-based learning applied to document
  recognition}.
\newblock \bibinfo{journal}{Proceedings of the IEEE}
  \bibinfo{year}{1998};\bibinfo{volume}{86}(\bibinfo{number}{11}):\bibinfo{pages}{2278--2324}.
\bibitem[{Li et~al.(2014)Li, Zhang, Chen and Smola}]{li2014efficient}
\bibinfo{author}{Li\xfnm[ M.]}, \bibinfo{author}{Zhang\xfnm[ T.]},
  \bibinfo{author}{Chen\xfnm[ Y.]}, \bibinfo{author}{Smola\xfnm[ A.J.]}.
\newblock \bibinfo{title}{Efficient mini-batch training for stochastic
  optimization}.
\newblock In: \bibinfo{booktitle}{Proceedings of the 20th ACM SIGKDD
  International Conference on Knowledge Discovery and Data Mining}.
  \bibinfo{organization}{ACM}; \bibinfo{year}{2014}. p.
  \bibinfo{pages}{661--670}.
\bibitem[{Linderman et~al.(2016)Linderman, Miller, Adams, Blei, Paninski and
  Johnson}]{linderman2016recurrent}
\bibinfo{author}{Linderman\xfnm[ S.W.]}, \bibinfo{author}{Miller\xfnm[ A.C.]},
  \bibinfo{author}{Adams\xfnm[ R.P.]}, \bibinfo{author}{Blei\xfnm[ D.M.]},
  \bibinfo{author}{Paninski\xfnm[ L.]}, \bibinfo{author}{Johnson\xfnm[ M.J.]}.
\newblock \bibinfo{title}{Recurrent switching linear dynamical systems}.
\newblock \bibinfo{journal}{arXiv preprint arXiv:161008466}
  \bibinfo{year}{2016};.
\bibitem[{Louizos et~al.(2017)Louizos, Shalit, Mooij, Sontag, Zemel and
  Welling}]{louizos2017causal}
\bibinfo{author}{Louizos\xfnm[ C.]}, \bibinfo{author}{Shalit\xfnm[ U.]},
  \bibinfo{author}{Mooij\xfnm[ J.M.]}, \bibinfo{author}{Sontag\xfnm[ D.]},
  \bibinfo{author}{Zemel\xfnm[ R.]}, \bibinfo{author}{Welling\xfnm[ M.]}.
\newblock \bibinfo{title}{Causal effect inference with deep latent-variable
  models}.
\newblock In: \bibinfo{booktitle}{Advances in Neural Information Processing
  Systems}. \bibinfo{year}{2017}. p. \bibinfo{pages}{6446--6456}.
\bibitem[{Maddison et~al.(2016)Maddison, Mnih and Teh}]{maddison2016concrete}
\bibinfo{author}{Maddison\xfnm[ C.J.]}, \bibinfo{author}{Mnih\xfnm[ A.]},
  \bibinfo{author}{Teh\xfnm[ Y.W.]}.
\newblock \bibinfo{title}{The {C}oncrete distribution: {A} continuous
  relaxation of discrete random variables}.
\newblock \bibinfo{journal}{CoRR}
  \bibinfo{year}{2016};\bibinfo{volume}{abs/1611.00712}.
\newblock \URLprefix \url{http://arxiv.org/abs/1611.00712}.
  \href{http://arxiv.org/abs/1611.00712}{\tt arXiv:1611.00712}.
\bibitem[{Masegosa et~al.(2017{\natexlab{a}})Masegosa, Nielsen, Langseth,
  Ramos-Lopez, Salmer{\'o}n and Madsen}]{masegosa2017bayesian}
\bibinfo{author}{Masegosa\xfnm[ A.]}, \bibinfo{author}{Nielsen\xfnm[ T.D.]},
  \bibinfo{author}{Langseth\xfnm[ H.]}, \bibinfo{author}{Ramos-Lopez\xfnm[
  D.]}, \bibinfo{author}{Salmer{\'o}n\xfnm[ A.]}, \bibinfo{author}{Madsen\xfnm[
  A.L.]}.
\newblock \bibinfo{title}{{B}ayesian models of data streams with hierarchical
  power priors}.
\newblock \bibinfo{journal}{arXiv preprint arXiv:170702293}
  \bibinfo{year}{2017}{\natexlab{a}};.
\bibitem[{Masegosa et~al.(2017{\natexlab{b}})Masegosa, Martinez, Langseth,
  Nielsen, Salmer{\'o}n, Ramos-L{\'o}pez and Madsen}]{masegosa2017scaling}
\bibinfo{author}{Masegosa\xfnm[ A.R.]}, \bibinfo{author}{Martinez\xfnm[ A.M.]},
  \bibinfo{author}{Langseth\xfnm[ H.]}, \bibinfo{author}{Nielsen\xfnm[ T.D.]},
  \bibinfo{author}{Salmer{\'o}n\xfnm[ A.]},
  \bibinfo{author}{Ramos-L{\'o}pez\xfnm[ D.]}, \bibinfo{author}{Madsen\xfnm[
  A.L.]}.
\newblock \bibinfo{title}{Scaling up {B}ayesian variational inference using
  distributed computing clusters}.
\newblock \bibinfo{journal}{International Journal of Approximate Reasoning}
  \bibinfo{year}{2017}{\natexlab{b}};\bibinfo{volume}{88}:\bibinfo{pages}{435--451}.
\bibitem[{Minka(2001)}]{minka2001expectation}
\bibinfo{author}{Minka\xfnm[ T.P.]}.
\newblock \bibinfo{title}{Expectation propagation for approximate {B}ayesian
  inference}.
\newblock In: \bibinfo{booktitle}{Proceedings of the Seventeenth conference on
  Uncertainty in artificial intelligence}. \bibinfo{organization}{Morgan
  Kaufmann Publishers}; \bibinfo{year}{2001}. p. \bibinfo{pages}{362--369}.
\bibitem[{Mnih and Gregor(2014)}]{mnih2014neural}
\bibinfo{author}{Mnih\xfnm[ A.]}, \bibinfo{author}{Gregor\xfnm[ K.]}.
\newblock \bibinfo{title}{Neural variational inference and learning in belief
  networks}.
\newblock \bibinfo{journal}{arXiv preprint arXiv:14020030}
  \bibinfo{year}{2014};.
\bibitem[{Mnih and Rezende(2016)}]{mnih2016variational}
\bibinfo{author}{Mnih\xfnm[ A.]}, \bibinfo{author}{Rezende\xfnm[ D.J.]}.
\newblock \bibinfo{title}{Variational inference for {M}onte {C}arlo
  objectives}.
\newblock \bibinfo{journal}{arXiv preprint arXiv:160206725}
  \bibinfo{year}{2016};.
\bibitem[{Murphy(2012)}]{murphy2012machine}
\bibinfo{author}{Murphy\xfnm[ K.P.]}.
\newblock \bibinfo{title}{Machine Learning: A Probabilistic Perspective}.
\newblock \bibinfo{publisher}{MIT Press}, \bibinfo{year}{2012}.
\bibitem[{Murphy et~al.(1999)Murphy, Weiss and Jordan}]{murphy1999loopy}
\bibinfo{author}{Murphy\xfnm[ K.P.]}, \bibinfo{author}{Weiss\xfnm[ Y.]},
  \bibinfo{author}{Jordan\xfnm[ M.I.]}.
\newblock \bibinfo{title}{Loopy belief propagation for approximate inference:
  An empirical study}.
\newblock In: \bibinfo{booktitle}{Proceedings of the Fifteenth conference on
  Uncertainty in artificial intelligence}. \bibinfo{organization}{Morgan
  Kaufmann Publishers}; \bibinfo{year}{1999}. p. \bibinfo{pages}{467--475}.
\bibitem[{Paszke et~al.(2017)Paszke, Gross, Chintala, Chanan, Yang, DeVito,
  Lin, Desmaison, Antiga and Lerer}]{paszke2017automatic}
\bibinfo{author}{Paszke\xfnm[ A.]}, \bibinfo{author}{Gross\xfnm[ S.]},
  \bibinfo{author}{Chintala\xfnm[ S.]}, \bibinfo{author}{Chanan\xfnm[ G.]},
  \bibinfo{author}{Yang\xfnm[ E.]}, \bibinfo{author}{DeVito\xfnm[ Z.]},
  \bibinfo{author}{Lin\xfnm[ Z.]}, \bibinfo{author}{Desmaison\xfnm[ A.]},
  \bibinfo{author}{Antiga\xfnm[ L.]}, \bibinfo{author}{Lerer\xfnm[ A.]}.
\newblock \bibinfo{title}{Automatic differentiation in {PyTorch}}.
\newblock \bibinfo{year}{2017}.
\newblock \bibinfo{note}{NIPS AutoDiff Workshop}.
\bibitem[{Pearl(1988)}]{Pearl88}
\bibinfo{author}{Pearl\xfnm[ J.]}.
\newblock \bibinfo{title}{Probabilistic Reasoning in Intelligent Systems:
  {N}etworks of Plausible Inference}.
\newblock \bibinfo{address}{San Mateo, CA.}: \bibinfo{publisher}{Morgan
  Kaufmann Publishers}, \bibinfo{year}{1988}.
\bibitem[{Pless and Souvenir(2009)}]{pless2009survey}
\bibinfo{author}{Pless\xfnm[ R.]}, \bibinfo{author}{Souvenir\xfnm[ R.]}.
\newblock \bibinfo{title}{A survey of manifold learning for images}.
\newblock \bibinfo{journal}{IPSJ Transactions on Computer Vision and
  Applications}
  \bibinfo{year}{2009};\bibinfo{volume}{1}:\bibinfo{pages}{83--94}.
\bibitem[{Plummer(2003)}]{plummer2003jags}
\bibinfo{author}{Plummer\xfnm[ M.]}.
\newblock \bibinfo{title}{{JAGS}: A program for analysis of {B}ayesian
  graphical models using {G}ibbs sampling}.
\newblock In: \bibinfo{booktitle}{Proceedings of the 3rd international workshop
  on distributed statistical computing}. \bibinfo{organization}{Vienna,
  Austria}; volume \bibinfo{volume}{124}; \bibinfo{year}{2003}. .
\bibitem[{Pritchard et~al.(2000)Pritchard, Stephens and
  Donnelly}]{pritchard2000inference}
\bibinfo{author}{Pritchard\xfnm[ J.K.]}, \bibinfo{author}{Stephens\xfnm[ M.]},
  \bibinfo{author}{Donnelly\xfnm[ P.]}.
\newblock \bibinfo{title}{Inference of population structure using multilocus
  genotype data}.
\newblock \bibinfo{journal}{Genetics}
  \bibinfo{year}{2000};\bibinfo{volume}{155}(\bibinfo{number}{2}):\bibinfo{pages}{945--959}.
\bibitem[{Pu et~al.(2016)Pu, Gan, Henao, Yuan, Li, Stevens and
  Carin}]{pu2016variational}
\bibinfo{author}{Pu\xfnm[ Y.]}, \bibinfo{author}{Gan\xfnm[ Z.]},
  \bibinfo{author}{Henao\xfnm[ R.]}, \bibinfo{author}{Yuan\xfnm[ X.]},
  \bibinfo{author}{Li\xfnm[ C.]}, \bibinfo{author}{Stevens\xfnm[ A.]},
  \bibinfo{author}{Carin\xfnm[ L.]}.
\newblock \bibinfo{title}{Variational autoencoder for deep learning of images,
  labels and captions}.
\newblock In: \bibinfo{booktitle}{Advances in Neural Information Processing
  Systems}. \bibinfo{year}{2016}. p. \bibinfo{pages}{2352--2360}.
\bibitem[{Ranganath et~al.(2014)Ranganath, Gerrish and
  Blei}]{ranganath2014black}
\bibinfo{author}{Ranganath\xfnm[ R.]}, \bibinfo{author}{Gerrish\xfnm[ S.]},
  \bibinfo{author}{Blei\xfnm[ D.]}.
\newblock \bibinfo{title}{Black box variational inference}.
\newblock In: \bibinfo{booktitle}{Artificial Intelligence and Statistics}.
  \bibinfo{year}{2014}. p. \bibinfo{pages}{814--822}.
\bibitem[{Rezende et~al.(2014)Rezende, Mohamed and
  Wierstra}]{rezende2014stochastic}
\bibinfo{author}{Rezende\xfnm[ D.J.]}, \bibinfo{author}{Mohamed\xfnm[ S.]},
  \bibinfo{author}{Wierstra\xfnm[ D.]}.
\newblock \bibinfo{title}{Stochastic backpropagation and approximate inference
  in deep generative models}.
\newblock \bibinfo{journal}{arXiv preprint arXiv:14014082}
  \bibinfo{year}{2014};.
\bibitem[{Robbins and Monro(1951)}]{robbins1951stochastic}
\bibinfo{author}{Robbins\xfnm[ H.]}, \bibinfo{author}{Monro\xfnm[ S.]}.
\newblock \bibinfo{title}{A stochastic approximation method}.
\newblock \bibinfo{journal}{The Annals of Mathematical Statistics}
  \bibinfo{year}{1951};\bibinfo{volume}{22}(\bibinfo{number}{3}):\bibinfo{pages}{400--407}.
\bibitem[{Ruiz et~al.(2016)Ruiz, Titsias and Blei}]{ruiz2016generalized}
\bibinfo{author}{Ruiz\xfnm[ F.]}, \bibinfo{author}{Titsias\xfnm[ M.]},
  \bibinfo{author}{Blei\xfnm[ D.]}.
\newblock \bibinfo{title}{The generalized reparameterization gradient}.
\newblock In: \bibinfo{booktitle}{Advances in Neural Information Processing
  Systems}. \bibinfo{year}{2016}. p. \bibinfo{pages}{460--468}.
\bibitem[{Russell and Norvig(2016)}]{russell2016artificial}
\bibinfo{author}{Russell\xfnm[ S.J.]}, \bibinfo{author}{Norvig\xfnm[ P.]}.
\newblock \bibinfo{title}{Artificial Intelligence: A Modern Approach}.
\newblock \bibinfo{publisher}{Pearson}, \bibinfo{year}{2016}.
\bibitem[{Salakhutdinov(2015)}]{salakhutdinov2015learning}
\bibinfo{author}{Salakhutdinov\xfnm[ R.]}.
\newblock \bibinfo{title}{Learning deep generative models}.
\newblock \bibinfo{journal}{Annual Review of Statistics and Its Application}
  \bibinfo{year}{2015};\bibinfo{volume}{2}:\bibinfo{pages}{361--385}.
\bibitem[{Salmer{\'o}n et~al.(2000)Salmer{\'o}n, Cano and
  Moral}]{salmeron2000importance}
\bibinfo{author}{Salmer{\'o}n\xfnm[ A.]}, \bibinfo{author}{Cano\xfnm[ A.]},
  \bibinfo{author}{Moral\xfnm[ S.]}.
\newblock \bibinfo{title}{Importance sampling in {B}ayesian networks using
  probability trees}.
\newblock \bibinfo{journal}{Computational Statistics \& Data Analysis}
  \bibinfo{year}{2000};\bibinfo{volume}{34}(\bibinfo{number}{4}):\bibinfo{pages}{387--413}.
\bibitem[{Salvatier et~al.(2016)Salvatier, Wiecki and
  Fonnesbeck}]{salvatier2016probabilistic}
\bibinfo{author}{Salvatier\xfnm[ J.]}, \bibinfo{author}{Wiecki\xfnm[ T.V.]},
  \bibinfo{author}{Fonnesbeck\xfnm[ C.]}.
\newblock \bibinfo{title}{Probabilistic programming in {P}ython using {PyMC3}}.
\newblock \bibinfo{journal}{PeerJ Computer Science}
  \bibinfo{year}{2016};\bibinfo{volume}{2}:\bibinfo{pages}{e55}.
\bibitem[{Sch{\"o}lkopf et~al.(1998)Sch{\"o}lkopf, Smola and
  M{\"u}ller}]{scholkopf1998nonlinear}
\bibinfo{author}{Sch{\"o}lkopf\xfnm[ B.]}, \bibinfo{author}{Smola\xfnm[ A.]},
  \bibinfo{author}{M{\"u}ller\xfnm[ K.R.]}.
\newblock \bibinfo{title}{Nonlinear component analysis as a kernel eigenvalue
  problem}.
\newblock \bibinfo{journal}{Neural Computation}
  \bibinfo{year}{1998};\bibinfo{volume}{10}(\bibinfo{number}{5}):\bibinfo{pages}{1299--1319}.
\bibitem[{Schulman et~al.(2015)Schulman, Heess, Weber and
  Abbeel}]{schulman2015gradient}
\bibinfo{author}{Schulman\xfnm[ J.]}, \bibinfo{author}{Heess\xfnm[ N.]},
  \bibinfo{author}{Weber\xfnm[ T.]}, \bibinfo{author}{Abbeel\xfnm[ P.]}.
\newblock \bibinfo{title}{Gradient estimation using stochastic computation
  graphs}.
\newblock In: \bibinfo{booktitle}{Advances in Neural Information Processing
  Systems}. \bibinfo{year}{2015}. p. \bibinfo{pages}{3528--3536}.
\bibitem[{Semeniuta et~al.(2017)Semeniuta, Severyn and
  Barth}]{semeniuta2017hybrid}
\bibinfo{author}{Semeniuta\xfnm[ S.]}, \bibinfo{author}{Severyn\xfnm[ A.]},
  \bibinfo{author}{Barth\xfnm[ E.]}.
\newblock \bibinfo{title}{A hybrid convolutional variational autoencoder for
  text generation}.
\newblock \bibinfo{journal}{arXiv preprint arXiv:170202390}
  \bibinfo{year}{2017};.
\bibitem[{Sohn et~al.(2015)Sohn, Lee and Yan}]{sohn2015learning}
\bibinfo{author}{Sohn\xfnm[ K.]}, \bibinfo{author}{Lee\xfnm[ H.]},
  \bibinfo{author}{Yan\xfnm[ X.]}.
\newblock \bibinfo{title}{Learning structured output representation using deep
  conditional generative models}.
\newblock In: \bibinfo{booktitle}{Advances in Neural Information Processing
  Systems}. \bibinfo{year}{2015}. p. \bibinfo{pages}{3483--3491}.
\bibitem[{Tipping and Bishop(1999)}]{tipping1999probabilistic}
\bibinfo{author}{Tipping\xfnm[ M.E.]}, \bibinfo{author}{Bishop\xfnm[ C.M.]}.
\newblock \bibinfo{title}{Probabilistic principal component analysis}.
\newblock \bibinfo{journal}{Journal of the Royal Statistical Society: Series B
  (Statistical Methodology)}
  \bibinfo{year}{1999};\bibinfo{volume}{61}(\bibinfo{number}{3}):\bibinfo{pages}{611--622}.
\bibitem[{Titsias and L{\'a}zaro-Gredilla(2014)}]{titsias2014doubly}
\bibinfo{author}{Titsias\xfnm[ M.]}, \bibinfo{author}{L{\'a}zaro-Gredilla\xfnm[
  M.]}.
\newblock \bibinfo{title}{Doubly stochastic variational {B}ayes for
  non-conjugate inference}.
\newblock In: \bibinfo{booktitle}{International Conference on Machine
  Learning}. \bibinfo{year}{2014}. p. \bibinfo{pages}{1971--1979}.
\bibitem[{Tran et~al.(2018)Tran, Hoffman, Moore, Suter, Vasudevan and
  Radul}]{tran2018simple}
\bibinfo{author}{Tran\xfnm[ D.]}, \bibinfo{author}{Hoffman\xfnm[ M.W.]},
  \bibinfo{author}{Moore\xfnm[ D.]}, \bibinfo{author}{Suter\xfnm[ C.]},
  \bibinfo{author}{Vasudevan\xfnm[ S.]}, \bibinfo{author}{Radul\xfnm[ A.]}.
\newblock \bibinfo{title}{Simple, distributed, and accelerated probabilistic
  programming}.
\newblock In: \bibinfo{booktitle}{Advances in Neural Information Processing
  Systems}. \bibinfo{year}{2018}. p. \bibinfo{pages}{7608--7619}.
\bibitem[{Tran et~al.(2016)Tran, Kucukelbir, Dieng, Rudolph, Liang and
  Blei}]{tran2016edward}
\bibinfo{author}{Tran\xfnm[ D.]}, \bibinfo{author}{Kucukelbir\xfnm[ A.]},
  \bibinfo{author}{Dieng\xfnm[ A.B.]}, \bibinfo{author}{Rudolph\xfnm[ M.]},
  \bibinfo{author}{Liang\xfnm[ D.]}, \bibinfo{author}{Blei\xfnm[ D.M.]}.
\newblock \bibinfo{title}{Edward: A library for probabilistic modeling,
  inference, and criticism}.
\newblock \bibinfo{journal}{arXiv preprint arXiv:161009787}
  \bibinfo{year}{2016};.
\bibitem[{Tucker et~al.(2017)Tucker, Mnih, Maddison, Lawson and
  Sohl-Dickstein}]{tucker2017rebar}
\bibinfo{author}{Tucker\xfnm[ G.]}, \bibinfo{author}{Mnih\xfnm[ A.]},
  \bibinfo{author}{Maddison\xfnm[ C.J.]}, \bibinfo{author}{Lawson\xfnm[ J.]},
  \bibinfo{author}{Sohl-Dickstein\xfnm[ J.]}.
\newblock \bibinfo{title}{Rebar: Low-variance, unbiased gradient estimates for
  discrete latent variable models}.
\newblock In: \bibinfo{booktitle}{Advances in Neural Information Processing
  Systems}. \bibinfo{year}{2017}. p. \bibinfo{pages}{2627--2636}.
\bibitem[{Wainwright and Jordan(2008)}]{wainwright2008graphical}
\bibinfo{author}{Wainwright\xfnm[ M.J.]}, \bibinfo{author}{Jordan\xfnm[ M.I.]}.
\newblock \bibinfo{title}{Graphical models, exponential families, and
  variational inference}.
\newblock \bibinfo{journal}{Foundations and Trends{\textregistered} in Machine
  Learning}
  \bibinfo{year}{2008};\bibinfo{volume}{1}(\bibinfo{number}{1--2}):\bibinfo{pages}{1--305}.
\bibitem[{Williams(1992)}]{williams1992simple}
\bibinfo{author}{Williams\xfnm[ R.J.]}.
\newblock \bibinfo{title}{Simple statistical gradient-following algorithms for
  connectionist reinforcement learning}.
\newblock \bibinfo{journal}{Machine Learning}
  \bibinfo{year}{1992};\bibinfo{volume}{8}(\bibinfo{number}{3-4}):\bibinfo{pages}{229--256}.
\bibitem[{Wingate and Weber(2013)}]{wingate2013automated}
\bibinfo{author}{Wingate\xfnm[ D.]}, \bibinfo{author}{Weber\xfnm[ T.]}.
\newblock \bibinfo{title}{Automated variational inference in probabilistic
  programming}.
\newblock \bibinfo{journal}{arXiv preprint arXiv:13011299}
  \bibinfo{year}{2013};.
\bibitem[{Winn and Bishop(2005)}]{WinnBishop05}
\bibinfo{author}{Winn\xfnm[ J.M.]}, \bibinfo{author}{Bishop\xfnm[ C.M.]}.
\newblock \bibinfo{title}{Variational message passing}.
\newblock \bibinfo{journal}{Journal of Machine Learning Research}
  \bibinfo{year}{2005};\bibinfo{volume}{6}:\bibinfo{pages}{661--694}.
\bibitem[{Xie et~al.(2016)Xie, Girshick and Farhadi}]{xie2016unsupervised}
\bibinfo{author}{Xie\xfnm[ J.]}, \bibinfo{author}{Girshick\xfnm[ R.]},
  \bibinfo{author}{Farhadi\xfnm[ A.]}.
\newblock \bibinfo{title}{Unsupervised deep embedding for clustering analysis}.
\newblock In: \bibinfo{booktitle}{International Conference on Machine
  Learning}. \bibinfo{year}{2016}. p. \bibinfo{pages}{478--487}.
\bibitem[{Zhang et~al.(2018)Zhang, {B\"{u}tepage}, {Kjellstr\"{o}m} and
  Mandt}]{zhang2018advances}
\bibinfo{author}{Zhang\xfnm[ C.]}, \bibinfo{author}{{B\"{u}tepage}\xfnm[ J.]},
  \bibinfo{author}{{Kjellstr\"{o}m}\xfnm[ H.]}, \bibinfo{author}{Mandt\xfnm[
  S.]}.
\newblock \bibinfo{title}{Advances in variational inference}.
\newblock \bibinfo{journal}{IEEE Transactions on Pattern Analysis and Machine
  Intelligence}
  \bibinfo{year}{2018};\bibinfo{volume}{41}:\bibinfo{pages}{2008--2026}.
\bibitem[{Zhou et~al.(2015)Zhou, Cong and Chen}]{zhou2015poisson}
\bibinfo{author}{Zhou\xfnm[ M.]}, \bibinfo{author}{Cong\xfnm[ Y.]},
  \bibinfo{author}{Chen\xfnm[ B.]}.
\newblock \bibinfo{title}{The {P}oisson {G}amma belief network}.
\newblock In: \bibinfo{booktitle}{Advances in Neural Information Processing
  Systems}. \bibinfo{year}{2015}. p. \bibinfo{pages}{3043--3051}.

\end{thebibliography}
